\documentclass[acmtog,screen,nonacm]{acmart}
\usepackage{stackengine}
\usepackage{colortbl}
\usepackage{xcolor}
\usepackage{booktabs}
\usepackage{multirow}
\usepackage{subcaption}
\usepackage{diagbox}

\usepackage{xcolor}
\definecolor{myred}{rgb}{0.8,0,0}
\definecolor{mygreen}{rgb}{0,0.8,0}
\usepackage{pifont}
\newcommand{\good}{{\color{mygreen}\ding{51}}}

\definecolor{red}{rgb}{1,0.6,0.6}
\definecolor{orange}{rgb}{1,0.8,0.6}
\definecolor{yellow}{rgb}{1,1,0.6}

\usepackage{soul}
\definecolor{junglegreen}{rgb}{0.02, 1.0, 0.27}

\definecolor{mypurple}{rgb}{ 0.174, 0.190, 0.876}

\newcommand{\titleabr}{\textit{UMA} }

\newcommand{\fst}[1]{\cellcolor{red}{#1}}
\newcommand{\snd}[1]{\cellcolor{orange}{#1}}
\newcommand{\trd}[1]{\cellcolor{yellow}{#1}}

\newcommand{\ffst}[1]{\colorbox{red}{#1}}
\newcommand{\fsnd}[1]{\colorbox{orange}{#1}}
\newcommand{\ftrd}[1]{\colorbox{yellow}{#1}}

\begin{document}

\title{UMA: Ultra-detailed Human Avatars via Multi-level Surface Alignment}

\author{Heming Zhu}
\email{hezhu@mpi-inf.mpg.de}

\affiliation{%
	\institution{Max Planck Institute for Informatics, Saarland Informatics Campus}
	\country{Germany}
}

\author{Guoxing Sun}
\email{gsun@mpi-inf.mpg.de}

\affiliation{%
	\institution{Max Planck Institute for Informatics, Saarland Informatics Campus}
	\country{Germany}
}

\author{Christian Theobalt}
\email{theobalt@mpi-inf.mpg.de}

\affiliation{%
	\institution{Max Planck Institute for Informatics, Saarland Informatics Campus and Saarbrücken Research Center for Visual Computing, Interaction and AI}
	\country{Germany}
}

\author{Marc Habermann}
\email{mhaberma@mpi-inf.mpg.de}

\affiliation{%
	\institution{Max Planck Institute for Informatics, Saarland Informatics Campus and Saarbrücken Research Center for Visual Computing, Interaction and AI}
	\country{Germany}
}

%
%
\begin{abstract}
Learning an animatable and clothed human avatar model with vivid dynamics and photorealistic appearance from multi-view videos is an important foundational research problem in computer graphics and vision.
Fueled by recent advances in implicit representations, the quality of the animatable avatars has achieved an unprecedented level by attaching the implicit representation to drivable human template meshes.
However, they usually fail to preserve highest level of detail, e.g., fine textures and yarn-level patterns, particularly apparent when the virtual camera is zoomed in and when rendering at 4K resolution and higher. 
We argue that this limitation stems from inaccurate surface tracking, specifically, depth misalignment and surface drift between character geometry and the ground truth surface, which forces the detailed appearance model to compensate for geometric errors.
To address this, we adopt a latent deformation model and supervise the 3D deformation of the animatable character using guidance from foundational 2D video point trackers, which offer improved robustness to shading and surface variations, and are less prone to local minima than differentiable rendering. 
To mitigate the drift over time and lack of 3D awareness of 2D point trackers, we introduce a cascaded training strategy that generates consistent 3D point tracks by anchoring point tracks to the rendered avatar, which ultimately supervise our avatar at vertex and texel level.
Furthermore, a lightweight Gaussian texture super-resolution module is employed to reconstruct challenging appearance details and micro-level structures using localized information.
To validate the effectiveness of our approach, we introduce a novel dataset comprising five multi-view video sequences, each over 10 minutes in duration, captured using 40 calibrated 6K-resolution cameras, featuring subjects dressed in clothing with challenging texture patterns and wrinkle deformations.
Our approach demonstrates significantly improved performance in rendering quality and geometric accuracy over the prior state of the art.
\end{abstract}
%
%



%
%
\begin{teaserfigure}
    \centering
    \includegraphics[width=\linewidth]{image/teaser_ver_1.jpg}
    \vspace{-10pt}
    \caption{
    Given skeletal poses and a virtual camera, \titleabr renders ultra-detailed clothed human appearance and synthesizes high-fidelity geometry.
    Notably, \titleabr enables users to \textbf{digitally zoom in}, allowing close inspection of texture details or even fine yarn-level patterns.
    Additionally, we introduce a new dataset featuring multi-view 6K video recordings, capturing subjects wearing clothing with challenging texture patterns and rich dynamics.
    The reconstructed animatable avatars can serve as a strong foundation for downstream applications, e.g., motion editing, pose re-targeting, and texture editing.
    The fidelity of the reconstructed avatars makes them particularly suitable for virtual and mixed reality, where users can closely observe fine-grained appearance details.
    }
    \label{fig:teaser}
\end{teaserfigure}
%
%
\maketitle

%
%
\section{Introduction} \label{sec:intro}
%
%
Creating photorealistic, animatable full-body humans -- i.e. a model that maps skeletal motion to deforming geometry and surface appearance -- is a longstanding and challenging problem in computer graphics and vision. 
However, digitizing high-quality virtual doubles while preserving clothing dynamics and rendering them in novel poses typically requires substantial manual efforts from skilled artists~\cite{wang2024survey}. 
Therefore, automating the digitization of photorealistic virtual humans by learning from multi-view videos have consistently emerged as highly active research topics. 
The ultimate goal is to replicate the finest appearance details and dynamic surface deformations just like those in real-captured videos.
%
%
\par
Driven by recent advances in neural 3D representations ~\cite{su2021nerf, kerbl20233d}, 
the fidelity of animatable clothed human avatars~\cite{li2024animatable,habermann2023hdhumans} has significantly improved.
They typically attach neural representations onto a drivable human skeleton~\cite{li2022tava}, a parametric human body model~\cite{ARAH}, or a deformable template mesh~\cite{Pang_2024_CVPR}.
However, these methods still fall short in \textbf{capturing and synthesizing high-frequency details}, e.g., fine texture patterns and detailed geometry.
We argue that it is primarily due to imprecise surface tracking during training as many approaches either use body models~\cite{ARAH} (not accounting for the surface deformations at all) or deformable templates whose deformations are not correctly tracked~\cite{habermann2023hdhumans,habermann2021}, which leads to conflicting supervision across views and frames.
This work attempts to solve the inaccuracy in the template tracking -- more specifically, \textbf{the misalignment in depth} and \textbf{drifting on the surface} -- by jointly considering multi-level surface registration and neural human rendering, resulting in noticeable improvements in recovering high-quality geometry and intricate appearance details.
%
%
\par
The \textbf{misalignment in depth} denotes the discrepancies between the learned and the ground-truth geometry along the camera view direction. We found the depth misalignment partly arises due to fact that the skeletal motion alone is an insufficient input conditioning to learn surface deformations, i.e., clothing dynamics, since these are inherently stochastic and subject to other factors such as rest state. 
Interestingly, this stochasticity can be mostly observed and learned from longer training videos, which prior work~\cite{ARAH,li2022tava} often avoids or their results show a significant quality degradation when the number of training frames is increased. 
To mitigate this issue, we adopt a per-frame latent code to a coarse geometry network accounting for the inherent one-to-many mapping ambiguity~\cite{liu2021neural} between the surface dynamics and skeletal motion during training. 
This effectively builds a subspace for large-scale clothing dynamics, which better captures the cloth deformations present in the real video, and facilitates the learning of surface correspondence.
%
%
\par
Moreover, prior works often supervise the learned animatable template on 3D point-clouds~\cite{shetty2023holoported,habermann2021,habermann2023hdhumans}, which suffers from \textbf{drifting on the surface} across frames, due to the missing surface correspondence.
To establish correspondence between the learned geometry and the ground truth surface observed in the multi-view video, some works~\cite{habermann2021} leverage differentiable rendering for supervision, which is known to get stuck in local minima due to the shading variations and the complicate clothing dynamics.
Instead, we leverage a foundational point tracker~\cite{karaev2024cotracker}, which is originally designed to compute correspondences across video frames.
However, directly tracking correspondence over long videos remains challenging due to the inevitable accumulation of drift over time.
To this end, we propose an avatar-guided point tracking strategy, which estimates correspondence between the template mesh and multi-view videos, effectively supervising the 3D surface dynamics.
Moreover, we propose a visibility-aware, cascaded filtering strategy based on the animatable template to aggregate multi-view 2D correspondences and refine them into reliable 3D correspondences, which serve as a more precise supervision for the animatable clothed human geometry at each training stage.

Although our method effectively resolves depth misalignment and surface drifting, achieving precise alignment between the dynamic surface of the animatable character and the ground-truth surface at the depth, vertex, and texel levels. 
However, it may still struggle to recover the most intricate appearance details, such as fine yarn patterns, due to the limited number of 3D Gaussian splats.
To this end, we further introduce a lightweight super-resolution module that up-samples the dynamic Gaussian splats with minimal computational overhead.

%
%
\par
We found that most existing multi-view datasets of clothed humans are limited by image resolution ~\cite{habermann2021deeper,peng2021neuralbody}, lack of challenging texture patterns ~\cite{li2024animatable}, and restricted skeletal pose diversity ~\cite{peng2021neuralbody}.
Towards better benchmarking, we present a novel dataset featuring five clothed human subjects, each recorded for ten minutes at 6K resolution using 40 calibrated cameras greatly sampling skeletal pose and induced surface dynamics. 
%
%
Our contributions are summarized as follows:
\begin{itemize}
\item A novel method for learning animatable avatars that capture and synthesize highest level of visual and geometric details.
\item A latent conditioning to account for stochastic geometry dynamics that can not be solely modeled by skeletal motion.
\item Establishing multi-level surface correspondence on the human surface using our proposed avatar-guided point tracking with off-the-shelf foundational 2D point trackers.
\item A benchmark dataset consisting of multi-view videos at an unprecedented resolution of 6k capturing humans wearing garments with intricate textures and rich surface dynamics.
\end{itemize}
Our qualitative and quantitative results (see Fig.~\ref{fig:teaser}) demonstrate a clear improvement over prior work in terms of detail recovery, dynamics capture, and overall visual appearance.
%
%
%
%
\section{Related Work} \label{sec:rw}
Our work focuses on animatable human rendering and geometry generation, where the model takes solely skeletal motions as input at test time. 
We do not cover methods related to replay~\cite{peng2021neuralbody,Lombardi2021MVP,wang2020learning,weng_humannerf_2022_cvpr,isik2023humanrf,xu2024representing,jiang2025reperformer}, reconstruction~\cite{xiang2021modeling,alldieck18b,alldieck19,habermann2020deepcap,habermann2019livecap,xiu2023econ,zhang2024sifu,zheng2025gstar,zhu2022registering}, or image-based free-view rendering~\cite{kwon2021neural, wang2021ibrnet,Remelli2022TexelAligned,shetty2023holoported,sun2025real}.
In the following sections, we review related works on animatable avatars, categorized by their shape representations, namely, mesh-based, implicit-based, and point-based approaches.
%
%
\par \noindent\textbf{Mesh-based Approaches.}
Textured meshes are the most prevalent representation for modeling clothed human avatars due to their compatibility with existing rendering and animation pipelines.
Early approaches reconstruct and animate the person-specific textured template through physical simulation~\cite{stoll2010video, guan2012drape}, retrieval from a video database~\cite{xu2011video}, or interpolating within a texture stack~\cite{casas14, shysheya2019textured}.
More recently, some works \cite{bagautdinov2021driving, xiang2021modeling,xiang2022dressing} adopt neural networks to learn the motion-dependent texture of the human template mesh from the multi-view videos.
\citet{habermann2021} further models motion-dependent surface deformations using a learnable embedded graph ~\cite{embedded} and captures dynamic appearance through motion-aware texture maps.
MeshAvatar~\cite{chen2024meshavatar} estimates motion-dependent surface deformations using front-and-back feature maps and jointly learns material properties from multi-view videos in an end-to-end fashion.
However, bounded by the limited resolution of the template meshes, the mesh-based approaches usually fail to recover the fine-grained geometry and appearance details.
%
%
\par \noindent\textbf{Implicit-based Approaches.} 
To increase the representation's capacity for modeling detailed appearance and geometry, implicit-based approaches combine implicit fields with explicit shape proxies, i.e., virtual bones~\cite{li2022tava}, parametric body models~\cite{loper15,STAR:2020,SMPL-X:2019,TotalCapture2018}, or person-specific template meshes~\cite{habermann2021,habermann2020deepcap}.
To better model the pose-dependent appearance of humans, recent studies ~\cite{liu2021neural, peng2021animatable, xu2021hnerf, NNA, zheng2023avatarrex, kwon2023deliffas, habermann2023hdhumans} further introduce motion-aware residual deformations in the canonical space on top of the mesh template.
Neural Actor~\cite{liu2021neural} leverages the texture space of a parametric human body model to extract pose-aware features for inferring motion-dependent dynamics, but it struggles to represent humans wearing loose clothing.
Moreover, it usually requires seconds to render a frame, due to the extensive sampling needed and the large MLP used to parameterize the neural radiance field.
TriHuman~\cite{zhu2023trihuman} represents the animatable character using a triplane defined over the texture space of the person-specific template mesh, enabling real-time photorealistic rendering and high-quality surface geometry generation at test time.
However, prior approaches either overlook surface deformations that cannot be solely attributed to skeletal poses, or they suffer from surface drift, which evitable leads to blurred appearance details and over-smoothed geometry.
%
%
\par \noindent\textbf{Point-based Approaches.} 
Point clouds have long served as a powerful and widely adopted representation for human avatar modeling due to their high flexibility in terms of modeling geometric shapes.
SCALE~\cite{ma2021scale} and POP~\cite{ma2021power} learn non-rigid deformations of dynamic clothing by representing the clothed body as dense point clouds parameterized in the UV space of parametric body models~\cite{loper2015smpl}.
To address the discontinuities in UV parameterization, FITE~\cite{lin2022learning} extracts pose-dependent features from orthographic projective maps.
In recent years, 3D Gaussian Splatting~\cite{kerbl20233d} has attracted considerable attention for its capability to generate high-quality renderings in real time, making it a popular choice for animatable clothed human avatars.
GART~\cite{lei2023gart}, 3DGS-Avatar~\cite{qian20233dgs}, GauHuman~\cite{hu2023gauhuman}, and HUGS~\cite{kocabas2023hugs} learn a canonicalized clothed human body represented with 3D Gaussian Splats from monocular videos and animate it using linear blend skinning.
However, since the attributes of the 3D Gaussian Splats are predicted by MLPs, they struggle to capture high-frequency appearance details due to the inherent smoothness bias of MLPs.
Therefore, ASH~\cite{Pang_2024_CVPR}, GaussianAvatar~\cite{hu2023gaussianavatar}, and Animatable Gaussians~\cite{li2024animatable} learn animatable characters with motion-aware appearance by leveraging convolutional neural networks in the UV space or in the orthographic projection space.
While achieving notable improvements in rendering quality, they still fall short in capturing the finest level of appearance details and reconstructing coherent, detailed surfaces, due to surface drift.
PhysAvatar~\cite{PhysAavatar24} tracks the human and clothing surface by initializing 3DGS on the template mesh and supervising the deformation with a photometric loss. However, since the color of the Gaussians is optimizable over time, the geometric error can be compensated with color changes, resulting in wrong correspondences and, therefore, blurred appearance details.
Moreover, generating the simulation ready clothing requires immense efforts from experts to resolve the collisions, and drape the garments on the body template at the initialization phase.
In contrast, by establishing accurate correspondences between the generated clothed human surface and the multi-view video using 2D point tracks, our approach captures fine-grained appearance details and reconstructs detailed as well as space-time coherent surface dynamics without manual intervention from the artists.
%
%
%
%
\section{Methodology} \label{sec:method}
%
%
%
\begin{figure*}[h]
    \centering
    \includegraphics[width=\linewidth]{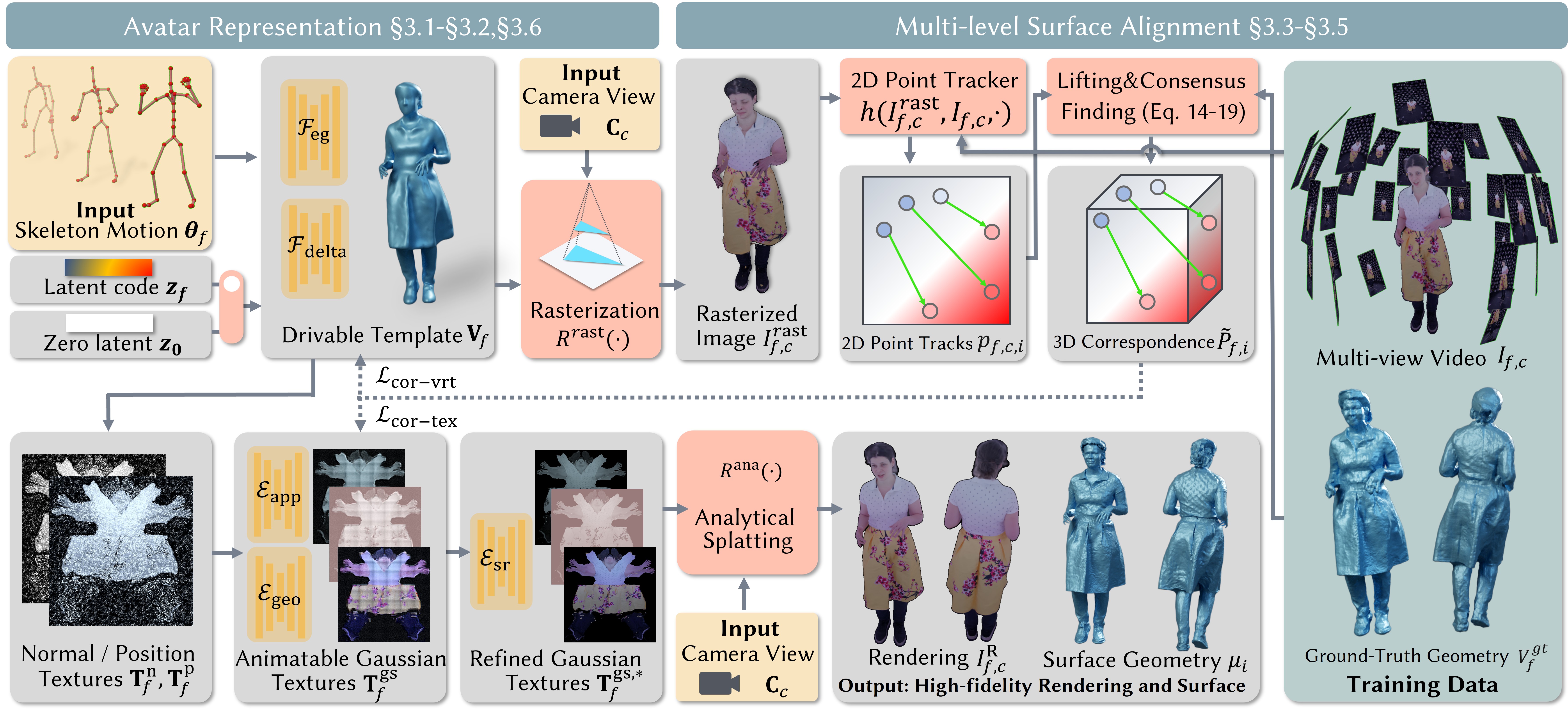}
    \vspace{-2em}
    \caption{
    \textbf{Overview.} \emph{UMA}, takes skeletal motion and the camera view as input and generates high-fidelity geometry and appearance. 
    To enhance the fidelity of the reconstructed human appearance and geometry, we tackle the problem from two key perspectives: \textbf{avatar representation} and \textbf{multi-level surface alignment}.
    For \textbf{avatar representation}, to address the stochasticity of the clothing dynamics that cannot be modeled by the skeletal motions, we inject a learnable latent code $\mathbf{z}_f$(zero latent $\mathbf{z}_{0}$ for testing) the drivable template $\mathbf{V}_f$ (Sec.~\ref{subsec:surface}).
    A texel super resolution module $\mathcal{E}_\mathrm{sr}$ is adopted to densify the animatable gaussian textures(Sec.~\ref{subsec:srtexel}).
    For \textbf{multi-level surface alignment}, we supervise the surface geometry at both the vertex (Sec.~\ref{subsec:triangle}) and texel levels (Sec.~\ref{subsec:texel}) using novel supervision derived from a foundational 2D point tracker.
    Specifically, the 2D point tracks $\mathbf{P}_{f,c,i}$ between the rasterized and ground-truth images obtained from the tracker are lifted and aggregated into 3D correspondences $\tilde{\mathbf{P}}_{f,i}$ across multiple views using the drivable template $\mathbf{V}_f$.
    }
    \label{fig:overview}
\end{figure*}
%
%
%
We aim to learn a photorealistic clothed human avatar with ultra-detailed appearance and surface geometry from multi-view videos.
Specifically, we target at capturing skeletal motion-aware surface dynamics and preserving finest details such as clothing wrinkles and texture pattens, allowing users to zoom in closely without noticeable quality degradation.
To this end, we introduce \titleabr, \textbf{U}ltra-detailed animatable clothed human avatar with \textbf{M}ulti-level surface \textbf{A}lignment, which during inference takes skeletal motion and camera views as input and generates photorealistic renderings at a resolution of $1620 \times 3072$ and detailed geometry $18$ fps.
An overview of our approach is illustrated in Fig.~\ref{fig:overview}.
%
%
\par 
In the following section, we will first introduce the underlying representation for the photoreal avatar (Sec.~\ref{sec:preliminaries}) and the key technical challenge (Sec.~\ref{subsec:key_idea}).
Subsequently, we elaborate how \titleabr captures fine-grained appearance and geometry details by improving surface tracking in terms of alignment in depth (Sec.~\ref{subsec:surface}) as well as surface drift at vertex (Sec.~\ref{subsec:triangle}) and texel level (Sec.~\ref{subsec:texel}). 
Finally, we present the light-weight texel super-resolution module for modeling finest appearance details (Sec.~\ref{subsec:srtexel}).
The adopted supervision and trained modules for each stage are summarized in Tab.~\ref{tab:status}.
%
%
%
\subsection{Gaussian-based Avatar Representation} \label{sec:preliminaries}
%
%
%
\par \noindent \textbf{Data Assumptions.} We assume segmented multi-view videos $\mathbf{I}_{f,c} \in \mathbb{R}^{H \times W}$, where $f$ and $c$ denote the frame and camera indices, respectively.
$W$ and $H$ denotes the width and height of captured multi-view imagery.
Each frame $\mathbf{I}_{f,c}$ is annotated with the camera camera calibrations $\mathbf{C}_c$ and 3D skeletal pose $\boldsymbol{\theta}_f \in \mathbb{R}^D$ using a commercial markerless motion capture system~\cite{captury}, where $D$ denotes the degrees of freedom (DoFs) of the 3D skeletal pose.
The motion $\boldsymbol{\bar{\theta}}_{f} \in \mathbb{R}^{k \times D}$ is derived from skeletal poses of a sliding window ranging from frame $f - k + 1$ to frame $f$ where the root joint translation is normalized w.r.t. the $f$th frame.
Moreover, we adopt implicit surface reconstruction~\cite{neus2} to recover per-frame ground truth geometry $\mathbf{V}_{f}^{\mathrm{gt}}$ and render it into depth maps $\mathbf{N}_{f,c} \in \mathbb{R}^{H \times W}$ for all camera views.
Notably, the reconstructed ground truth surfaces $\mathbf{V}_{f}^{\mathrm{gt}}$ are per-frame reconstructions and therefore they \emph{lack surface correspondence} over time.
%
%
\par \noindent \textbf{Drivable Human Template Mesh.} 
We first define a drivable and deformable template mesh
%
\begin{align}
M(\boldsymbol{\bar{\theta}}_{f}) \label{eq:template}
&= f_\mathrm{dq}(f_\mathrm{eg}(\mathcal{F_\mathrm{eg}}(\bar{\boldsymbol{\theta}}_{f}),\bar{\mathbf{V}})+ \mathcal{F_\mathrm{delta}}(\bar{\boldsymbol{\theta}}_{f}),\boldsymbol{\theta}_f) \\
&= f_\mathrm{dq}(f_\mathrm{eg}(\mathbf{A}_f, \mathbf{T}_f,\bar{\mathbf{V}})+ \boldsymbol{\delta}_f,\boldsymbol{\theta}_f) \\
&=f_\mathrm{dq}(\bar{\mathbf{V}}_{f},\boldsymbol{\theta}_f) \\
&=\mathbf{V}_{f}
\end{align}
%
of a clothed human to model coarse-level geometry.
It takes the skeletal motion $\bar{\boldsymbol{\theta}}_{f}$ as input and regresses posed and non-rigidly deformed 3D vertices $\mathbf{V}_{f}$ of a person-specific template mesh $\bar{\mathbf{V}} \in \mathbb{R}^{V \times 3}$.
Precisely, to model the motion-aware deformation of the clothed human, we follow ~\citet{habermann2021} and first apply embedded deformation $f_\mathrm{eg}(\cdot)$~\cite{embedded} where deformation parameters are predicted by a graph convolutional neural network
\begin{equation}
\mathcal{F_\mathrm{eg}}(\bar{\boldsymbol{\theta}}_{f}) = \mathbf{A}_f, \mathbf{T}_f,
\end{equation}
where $\mathbf{A}_f \in \mathbb{R}^{V \times 4}$ and $\mathbf{T}_f \in \mathbb{R}^{V \times 4}$ denotes the translation and rotation quaternions for the embedded graph nodes in the canonical space. 
Further, per-vertex displacements $\boldsymbol{\delta}_f$ are predicted as a function of skeletal motion by a second network
\begin{equation}
\boldsymbol{\delta}_f = \mathcal{F_\mathrm{delta}}(\bar{\boldsymbol{\theta}}_{f}).
\end{equation}
Those deformations are applied onto the canonical mesh template $\bar{\mathbf{V}}$ from coarse to fine, i.e., first embedded deformations are applied followed by the per-vertex displacements.
Lastly, the canonical and non-rigidly deformed template $\bar{\mathbf{V}}_{f}$ is posed w.r.t. the skeletal pose ${\boldsymbol{\theta}}_{f}$ using Dual Quaternion skinning~\cite{kavan2007skinning} $f_\mathrm{dq}(\cdot)$.
%
%
\par \noindent \textbf{Training the Drivable Template.} 
The two networks, i.e., $\mathcal{F_\mathrm{eg}}$ and $\mathcal{F_\mathrm{delta}}$ can be trained by minimizing the loss
%
\begin{equation} \label{eq:loss_template}
\mathcal{L}_\mathrm{temp}(\mathbf{V}_f) = \mathcal{L}_\mathrm{cham}(\mathbf{V}_f, \mathbf{V}^\mathrm{gt}_f) + \mathcal{L}_\mathrm{spatial}(\mathbf{V}_f)
\end{equation}
%
for all frames, where the first term compares the posed and deformed template against the ground truth surface in terms of Chamfer distance while the second term 
%
\begin{equation}
    \mathcal{L}_\mathrm{spatial}(\mathbf{V}_f) = \mathcal{L}_\mathrm{lap}(\mathbf{V}_f) + \mathcal{L}_\mathrm{lapz}(\mathbf{V}_f) +\mathcal{L}_\mathrm{norm}(\mathbf{V}_f)
\end{equation}
%
is a combination of spatial regularization terms, namely, the mesh Laplacian loss $\mathcal{L}_\mathrm{lap}$, Laplacian smoothness term $\mathcal{L}_\mathrm{lapz}$, and face normal consistency loss $\mathcal{L}_\mathrm{norm}$.
We refer to the supplemental document for more details regarding the embedded deformation and the geometry regularization.
%
%
\par \noindent \textbf{Motion-aware Animatable Gaussian Textures.} 
Building upon the drivable template mesh, we model the \emph{fine-grained} and motion-aware appearance and geometry of the clothed human using Gaussian textures $\mathbf{T}^{\mathrm{gs}}_f \in \mathbb{R}^{N\times62}$ ~\cite{Pang_2024_CVPR} in the template's UV space. 
Each of the $N$ texels covered by a triangle stores the parameters of a 3D Gaussian splat $ (\boldsymbol{\bar{\mu}}^{\mathrm{uv}}_i, \mathbf{\bar{d}}^{\mathrm{uv}}_i, \mathbf{q}^{\mathrm{uv}}_i, \mathbf{s}^{\mathrm{uv}}_i, \mathbf{\alpha}^{\mathrm{uv}}_i, \boldsymbol{\eta}^{\mathrm{uv}}_i)_f \in \mathbb{R}^{62}$ where $i$ denotes the $i$th texel.
Notably, the canonical Gaussian position $\boldsymbol{\bar{\mu}}_{\mathrm{uv},i}$ is derived from the non-rigidly deformed template mesh $\bar{\mathbf{V}}_{f}$ through barycentric interpolation.
To model the finer-level details, i.e., wrinkles, of the dynamic clothed human, a learnable motion-aware offset $\mathbf{\bar{d}}_{\mathrm{uv}, i}$ is applied to each Gaussian Splat in the canonical space.
Similar to the drivable template mesh, the Gaussian splats can be posed from the deformed canonical position $(\boldsymbol{\bar{\mu}}_i +\mathbf{\bar{d}}_{\mathrm{uv}, i}) $ to the position $\boldsymbol{\mu}_i$ in posed space through Dual Quaternion skinning ~\cite{kavan2007skinning}.
The remaining parameters denote the rotation quaternion $\mathbf{q}^{\mathrm{uv}}_i$, anisotropic scaling $\mathbf{s}^{\mathrm{uv}}_i$, opacity $\mathbf{\alpha}^{\mathrm{uv}}_i$, and spherical harmonics coefficients $\boldsymbol{\eta}^{\mathrm{uv}}_i$.
%
%
\par \noindent \textbf{Predicting Motion-aware Gaussians.} 
Thanks to the texel-based parameterization, the mapping between the skeletal motion $\boldsymbol{\bar{\theta}}_{f}$ and motion-aware Gaussian splats $\mathbf{T}^{\mathrm{gs}}_f \in \mathbb{R}^{N\times62}$ can be formulated as an image-to-image translation task~\cite{CycleGAN2017}.
Specifically, the skeletal motion $\boldsymbol{\bar{\theta}}_{f}$ is represented using the positional $\mathbf{T}^{\mathrm{p}}_f$ and normal textures $\mathbf{T}^{\mathrm{n}}_f$ of the posed and deformed template $\mathbf{V}_f$.
Then, two separate convolutional decoders~\cite{Pang_2024_CVPR} $\mathcal{E}_{\mathrm{geo}}(\mathbf{T}^{\mathrm{n}}_f, \mathbf{T}^{\mathrm{p}}_f)$ and $\mathcal{E}_{\mathrm{app}}(\mathbf{T}^{\mathrm{n}}_f, \mathbf{T}^{\mathrm{p}}_f)$ regress the geometry $ ( \mathbf{\bar{d}}^{\mathrm{uv}}_i, \mathbf{q}^{\mathrm{uv}}_i, \mathbf{s}^{\mathrm{uv}}_i, \mathbf{\alpha}^{\mathrm{uv}}_i)_f$ and appearance attributes $(\boldsymbol{\eta}^{\mathrm{uv}}_i)_f$ of the 3D Gaussian Splats, respectively.
%
%
\par \noindent \textbf{Splatting Gaussians to Image Space.} 
We adopt Analytical Splatting~\cite{liang2024analyticsplatting} 
%
\begin{equation}
 I^\mathrm{R}_{f,c} = R^\mathrm{ana}(\mathbf{T}^{\mathrm{gs}}_f, \mathbf{C}_c)
\end{equation}
%
to render the image $I^\mathrm{R}_{f,c}$ by integration over the actual pixel size during the splatting.
This provides improved anti-aliasing over the original 3DGS tile-based rasterizer, which is particularly useful when supervising on ultra-resolution data, i.e., 6K resolution, and when focusing on detail preservation -- the goal of this work.
\par 
\noindent \textbf{Training the Motion-aware Gaussians.} 
The Gaussian textures $\mathbf{T}^{\mathrm{gs}}_f$ are supervised on multi-view frames with a combined loss
%
\begin{equation} \label{eq:loss_gauss}
\mathcal{L}_\mathrm{gau}(I^\mathrm{R}_{f,c}) =
\mathcal{L}_{1}(I^\mathrm{R}_{f,c}, \mathbf{I}_{f,c}) + \mathcal{L}_\mathrm{ssim}(I^\mathrm{R}_{f,c}, \mathbf{I}_{f,c}), + \mathcal{L}_\mathrm{mrf}(I^\mathrm{R}_{f,c}, \mathbf{I}_{f,c}) 
\end{equation}
%
consisting of image-based $\mathcal{L}_{1}$, structural $\mathcal{L}_\mathrm{ssim}$, and perceptual losses $\mathcal{L}_\mathrm{mrf}$ ~\cite{wang2018image}.
%
%
\subsection{Key Technical Challenge} \label{subsec:key_idea}
While animatable Gaussian textures conditioned on the drivable template mesh can capture motion-aware appearance, they fail to capture and synthesize fine-grained appearance and geometry details for the following reasons (see also Fig.~\ref{fig:ablation}).
%
%
\par
\begin{figure}[t]
    \centering
    \includegraphics[width=\linewidth]{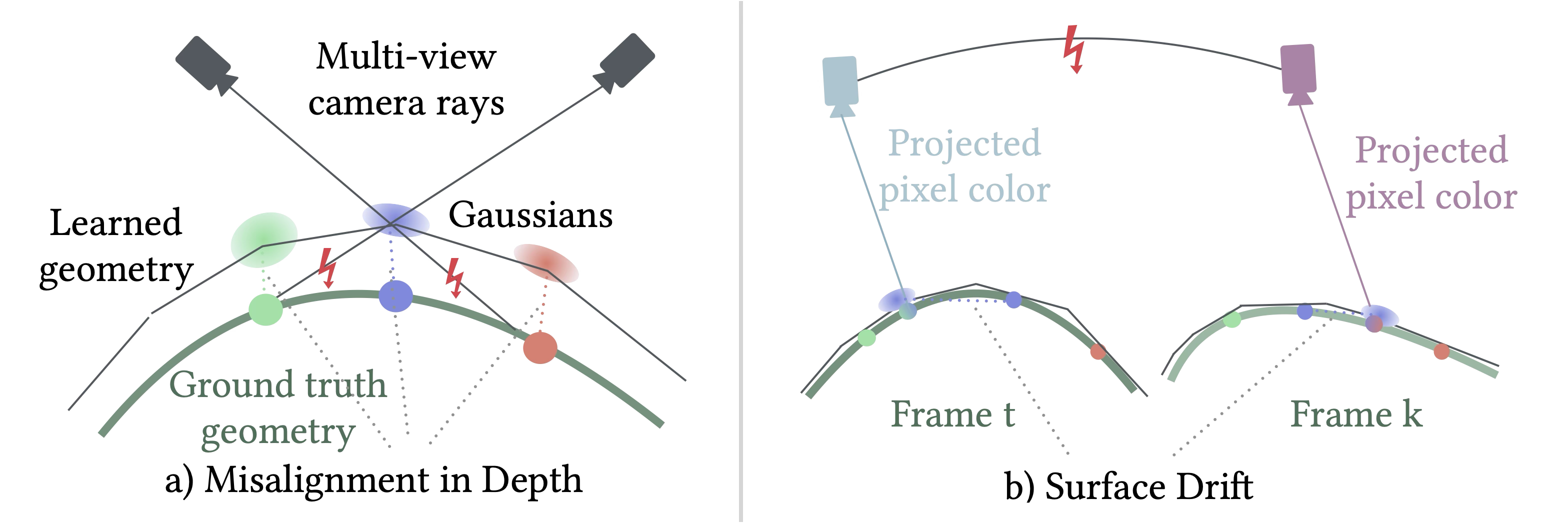}
    \vspace{-2em}
    \caption{
    \textbf{Key Technical Challenges.}
    (a) The depth misalignment leads to gradient conflicts when supervised with multi-view images, like the blue splat. 
    (b) Even if the depth misalignment is resolved, the surface drift between different frames results in gradient conflicts, resulting in averaged and blurry appearance. 
    }
    \vspace{-12pt}
    \label{fig:keyproblems}
\end{figure}

 \noindent \textbf{Misalignment in Depth.} 
In the context of animatable clothed human avatars, the highly diverse clothing dynamics cannot be fully explained by skeletal motions alone, as similar skeletal poses may result in significantly different surface deformations, often referred to as the one-to-many mapping issue~\cite{liu2021neural}. 
As a result, when using Eq.~\ref{eq:loss_template} to supervise the avatar's geometry (Eq.~\ref{eq:template}), the regression-based approach predicts an average surface that fits the multiple possible surface deformations in a least-squares sense due to this one-to-many mapping despite being supervised on accurate ground truth geometry. 
This manifests in misalignment in depth, which hinders accurate supervision of fine-grained appearance details, i.e., the motion-aware Gaussian Splats $\mathbf{T}^{\mathrm{gs}}_f$, since Gaussians may be wrongly projected onto training images, and therefore leads to blur or ghosting artifacts due to the conflicting supervision from different views (see Fig.~\ref{fig:keyproblems} a).
%
%
\par \noindent \textbf{On-surface Drift.} 
Even if depth misalignment is resolved, the ground-truth geometry obtained from implicit-based methods~\cite{neus2} still lacks temporal correspondence among frames \emph{and} correspondence between the template surface and the ground truth surface.
Thus, the second source of error is noticeable drift on the surface between the drivable template mesh and the ground-truth surface since the Chamfer loss can be minimized even if the correspondence is wrong.
For example in Fig.~\ref{fig:keyproblems} b), the blue Gaussian is never mapped onto the correct location (marked in blue).
Nonetheless, the Chamfer loss is minimized.
Even worse, the Gaussian is mapped onto different ground truth surface locations for different frames.
Thus, the rendering losses will assign different colors for different frames to the same Gaussian leading to a blurred average.
%
%
\par \noindent \textbf{Impact on Avatar Quality.} 
Such surface tracking errors directly translate into degradation of the avatar's quality since the network tries to compensate for the surface drift by changing colors dependent on the skeletal motion. 
This requires a lot of network capacity and, more importantly, it introduces stochastic effects that cannot be modeled by feed-forward architectures that are solely conditioned on motion. 
Thus, the network tends to predict the average color and over-smoothed surface deformations, causing the observed blur or, in other words, a reduced level of detail.
%
%
\subsection{Depth Alignment}  \label{subsec:surface}
%
%
%
%
%
%
To address the misalignment in depth caused by the one-to-many mapping between the skeletal motion and the surface geometry, we adopt a learnable per-frame latent 
%
\begin{equation} 
 \mathcal{H}_{\mathrm{lat}}(e(f/f_{\mathrm{max}})) = \mathbf{z}_f \in \mathbb{R}^{16}
\end{equation}
%
as an additional input alongside the skeletal motion $\boldsymbol{\bar{\theta}}_{f}$ for predicting the drivable template mesh $\bar{\mathbf{V}}_{f}$.
This latent accounts for the stochasticity and effectively establishes a one-to-one mapping.
Here, $f_\mathrm{max}$ denotes the maximum number of training frames, $e(\cdot)$ is a positional encoding~\cite{mildenhall2020nerf}, and $\mathcal{H}_{\mathrm{lat}}$ is a shallow MLP.
We then update our deformable mesh (Eq.~\ref{eq:template}) by concatenating the learnable per-frame latent code $\boldsymbol{z}_t$ channel-wise to the input graph nodes of, both, the embedded deformation network $\mathcal{F}_{\mathrm{eg}}(\bar{\boldsymbol{\theta}}_{f},\mathbf{z}_f)$ and the per-vertex deformation network $\mathcal{F}_{\mathrm{delta}}(\bar{\boldsymbol{\theta}}_{f},\mathbf{z}_f)$.
%
%
\par
We refer to the drivable template mesh as $M(\boldsymbol{\bar{\theta}}_{f}, \mathbf{z}_f)$ -- now also taking the learnable latent as input -- and then train the drivable human avatar as before using the losses defined in Eq.~\ref{eq:loss_template}.
At test time, we set $\mathbf{z}_f$ to $0$ to animate the character with novel motions, as it is intractable to predict the stochastic variations for unseen motions, and their influence on appearance and geometry is relatively minor compared to skeletal motion.

Moreover, to improve generalization to novel poses at the test time, inspired by ~\citet{li2024animatable}, we build a subspace $f_{\mathrm{pca}}$ of canonically deformed template meshes $\bar{\mathbf{V}}_{f}$ via Principal Component Analysis (PCA)~\cite{mackiewicz1993principal}.
To this end, the pose deformed template mesh under novel poses $\mathbf{V}^{\star}$ is computed through
%
\begin{align}
\mathbf{V}^{\star} &= f_\mathrm{dq}(f_{\mathrm{pca}}(\bar{\mathbf{V}}),\boldsymbol{\theta}_f) \\
&= f_\mathrm{dq}(\bar{\mathbf{V}}^{\star},\boldsymbol{\theta}_f)
\end{align}
%
where $\bar{\mathbf{V}}^{\star}$ is the canonical deformed template mesh in novel poses.
\par
As an design alternative to compensate for stochastic effects of the clothed human that cannot be modeled by skeleton motions, \citet{xiang2021modeling} introduce latent channels for both coarse and fine-level geometry and appearance, while also minimizing the mutual information between the posed template and the latent channels.
In contrast, we apply latent conditioning only to the coarse-level geometry, i.e., the template meshes $\bar{\mathbf{V}}_{f}$, whereas the finer-level geometry and appearance are conditioned solely on the coarse geometry without direct latent conditioning.
This decomposition effectively models the stochasticity of large-scale clothing dynamics through the latent-conditioned template mesh, enabling the preservation of fine geometric and appearance details in subsequent stages of animatable Gaussian texture learning, while mitigating artifacts and jittering for novel poses.
%
%
\subsection{Vertex-level Alignment}  \label{subsec:triangle}
Through the depth alignment (Sec.~\ref{subsec:surface}), the gap between the posed template mesh $\mathbf{V}_{f}$ and the ground truth surface $\mathbf{V}^\mathrm{gt}_{f}$ is significantly reduced by addressing the motion ambiguities.
However, since the ground truth reconstruction $\mathbf{V}^{\mathrm{gt}}_f$ inherently lacks correspondence over time, surface drift is still inevitable, which will result in blurred appearance and geometry, as discussed in Sec.~\ref{subsec:key_idea}.
Some prior works~\cite{habermann2021} propose to additionally supervise the geometric deformation with a differentiable image rendering loss to account for the drifting.
However, we found that such loss is easily stuck in local minima while not being robust to shading and change in lighting (see also Sec.~\ref{subsec:ablations}).
Instead, our solution builds up on recent advances on foundational 2D point tracking methods~\cite{karaev2024cotracker} that offer a promising alternative.
%
%
\par \noindent \textbf{2D Point Tracking.}
A 2D point tracker~\cite{karaev2024cotracker} estimates 2D point tracks between a pair of images $(\mathbf{I},\mathbf{I}')$, e.g., consecutive frames of a video.
More formally, it can be defined as the function
%
\begin{equation}
    h(\mathbf{I},\mathbf{I}',\mathbf{p})= \mathbf{p}' \in \mathbb{R}^2
\end{equation}
%
taking the image pair as well as a 2D image coordinate $\mathbf{p} \in \mathbb{R}^2$ (of image $I$) and estimates the corresponding matching point $\mathbf{p}'$ in image $I'$.
Since the point trackers are trained on large-scale real world datasets, they are typically robust to lighting changes and less likely to be trapped in local optima.
However, our goal is to carefully align our drivable template $M(\boldsymbol{\bar{\theta}}_{f}, \mathbf{z}_f)$ over a long multi-view sequence, which is not natively supported by the original video point trackers.
%
%
\par \noindent \textbf{Render-to-image 2D Point Tracks.}
Our key idea is to compute 2D correspondence between the rendering $\mathbf{I}^\mathrm{rast}_{f,c}$ of the deformed template $\mathbf{V}_f$ observed from camera $c$ using a rasterizer~\cite{Laine2020diffrast} $R^\mathrm{rast}(\mathbf{V}_f, \mathbf{T}_{\mathrm{0}}, \mathbf{C}_c)=\mathbf{I}^\mathrm{rast}_{f,c}$ and the ground truth training frame $\mathbf{I}_{f,c}$.
Notably, as the deformed template suffers from surface drift, the deformable mesh rendering and the ground truth do not perfectly align and the point tracks $h(\mathbf{I}^\mathrm{rast}_{f,c},I_{f,c},\mathbf{p})$ 
$\mathbf{p}_{f,c,i} \in \mathbb{R}^2$
will capture exactly this drift.
To acquire the static texture $\mathbf{T}_{\mathrm{0}}$ of the deformable mesh, we perform texture unprojection~\cite{shetty2023holoported} using the first frame of the multi-view video $\mathbf{I}_{0,c}$.
Then, we can compute the drift for each vertex $\mathbf{V}_{f,i} \in \mathbb{R}^3$, i.e., the 2D point tracks $\mathbf{p}_{f,c,i}$, in image space as 
%
\begin{align} 
\small 
\label{eq:vertex_correspondence}
    \mathbf{p}_{f,c,i} &= h(\mathbf{I}^\mathrm{rast}_{f,c},\mathbf{I}_{f,c},\mathbf{C}_c  \mathbf{V}_{f,i}) \\ 
    &= h(\mathbf{I}^\mathrm{rast}_{f,c},\mathbf{I}_{f,c},\mathbf{v}_{f,c,i}) 
\end{align}
by querying $h(\cdot,\cdot,\cdot)$ at the vertices' projected 2D position $\mathbf{v}_{c,f,i} \in \mathbb{R}^2$.
%
%

\par 

\noindent \textbf{3D Lifting and Consensus Finding.}
Next, we lift per-view 2D correspondences $\{\mathbf{p}_{f,c,i}\}_{c=1}^C$ of vertex $i$ to 3D correspondences $\{\mathbf{P}_{f,c,i}\}_{c=1}^C$ by querying depth maps followed by applying the inverse camera transformation: 
\begin{equation}
\mathbf{P}_{f,c,i} = \mathbf{C}^{-1}_c\mathbf{N}_{f,c}[\mathbf{p}_{f,c,i}],
\end{equation}
where $[\cdot]$ denotes bilinear interpolation.
To find consensus across views, we define a score function 
%
\begin{equation} \label{eq:scorefunction}
    s_{f,c,i} =  \frac{(\mathbf{V}_{f,i}-\mathbf{o}_{c})}{||\mathbf{V}_{f,i}-\mathbf{o}_{c}||}  \cdot -\mathbf{n}_i
\end{equation}
%
ranking views higher where the vertex normal $\mathbf{n}_i \in \mathbb{R}^3$ is parallel to camera ray defined by the cameras origin $\mathbf{o}_{c}$.
The optimal camera view 
%
\begin{equation} \label{eq:3D_vertex_correspondence}
    c^\mathrm{opt}_{f,i} = \arg\max_{c} (s_{f,c,i} * v_{f,c,i}),
\end{equation}
%
is the maximum of the score times the vertex' visibility $v_{f,c,i} \in \{0,1\}$ defining the final 3D correspondence point as 
\begin{equation}
\tilde{\mathbf{P}}_{f,i}=\mathbf{P}_{f,c^\mathrm{opt}_{f,i},i}.
\end{equation}
%
%
\par \noindent \textbf{Vertex Alignment Loss.}
Our template vertex alignment loss 
%
\begin{equation} \label{eq:loss_corr_templ}
\mathcal{L}_\mathrm{cor-vrt}(\mathbf{V}_{f}) = \sum_{f=1}^F \sum_{i \in \mathcal{V}_f} || \mathbf{V}_{f,i} - \tilde{\mathbf{P}}_{f,i} ||_2^2
\end{equation}
%
sums over all frames while only considering the set $\mathcal{V}_f$ of valid correspondences, i.e., pairs $(\mathbf{V}_{f,i}, \tilde{\mathbf{P}}_{f,i})$ whose distance is less than $3cm$.
%
%
\par \noindent \textbf{Refined Training of the Drivable Template.}
Our drift-aware training objective 
\begin{equation}
\mathcal{L}_\mathrm{template} + \mathcal{L}_\mathrm{cor-vrt}
\end{equation}
is now a combination of Eq.~\ref{eq:loss_template} and Eq.~\ref{eq:loss_corr_templ},
which is used to train our drivable and depth-aligned template $M(\boldsymbol{\bar{\theta}}_{f}, \mathbf{z}_f)$.
In practice, after optimizing the template mesh $\mathbf{V}_{f}$ via vertex-level alignment, we update the static texture $\mathbf{T}_{\mathrm{0}}$, recompute 2D correspondences using the (already) refined template, and perform a second round of vertex-level alignment.
This cascaded refinement improves the coherency and accuracy of the template tracking, as evidenced by the qualitative and quantitative results in Sec.~\ref{subsec:ablations}.
%
%
\subsection{Texel-level Alignment} \label{subsec:texel}
Through the depth and vertex-level alignment, we can now generate a drivable template mesh $M(\boldsymbol{\bar{\theta}}_{f}, \mathbf{z}_f)=\mathbf{V}_{f}$, which is closely attached to the ground truth surface $\mathbf{V}_{\mathrm{gt},f}$ and in correspondence over time.
Conditioned on this template $\mathbf{V}_f$, we can further train the detail dynamic human appearance represented with animatable Gaussian textures $\mathbf{T}^{\mathrm{gs}}_f$ by minimizing $\mathcal{L}_\mathrm{gau}$ (Eq.~\ref{eq:loss_gauss}).
However, similar to the template mesh vertices, the Gaussian texels may also drift on the surface hindering learning the highest frequency detail (see the discussion in Sec.~\ref{subsec:key_idea}).
Thus, we introduce an additional texel-level correspondence loss $\mathcal{L}_\mathrm{cor-tex}$, which serves as an additional regularization term for the position of the 3D Gaussian Splats $\boldsymbol{\mu}_i$:
\begin{equation} 
\mathcal{L}_\mathrm{cor-tex} = \sum_{f=1}^F \sum_{i \in \mathcal{V}^\mathrm{tex}_f} || \boldsymbol{\mu}_i -\tilde{\mathbf{P}}^\mathrm{tex}_{f,i}||_2^2,
\end{equation}
where $\tilde{\mathbf{P}}^\mathrm{tex}_{f,i} \in \mathbb{R}^2$ denotes the corresponding position for the $i$th texel on the ground truth surface.
The derivation of $\tilde{\mathbf{P}}^\mathrm{tex}_{f,i}$ is analogous to the one for vertices (Sec.~\ref{subsec:triangle}).
%
We can now even compensate for finest level drifts of the Gaussian textures by supervising them on a combined loss
\begin{equation}
\mathcal{L}_\mathrm{gau} + \mathcal{L}_\mathrm{cor-tex},
\end{equation}
further improving detail preservation (see Sec.~\ref{subsec:ablations}).
%
\subsection{Texel Super Resolution} \label{subsec:srtexel}
Increasing the resolution of the Gaussian Textures, i.e., the number of the 3D Gaussian Splats, could effectively improve the quality of modeling the fine structures, for example, the yarn patterns on the clothing.
However, directly tracking and learning high resolution Gaussian Textures is slow and memory extensive.
Therefore, rather than learn the highest resolution directly, we propose a light-weight Gaussian Texture super-resolution module $\mathcal{E}_\mathrm{sr}$ to further improve the fidelity of the Gaussian Texture.
The Gaussian texture super-resolution module takes the Gaussian Texture $\mathbf{T}^{\mathrm{gs}}_f$ from the geometry and appearance network, and produces the up-sampled Gaussian Texture $\mathbf{T}^{\mathrm{gs},*}_f$ represented with the residual w.r.t., the original textures:
\begin{equation}
\mathbf{T}^{\mathrm{gs},*}_f = \mathcal{E}_\mathrm{sr}(\mathbf{T}^{\mathrm{gs}}_f) + f_\mathrm{int}(\mathbf{T}^{\mathrm{gs}}_f)
\end{equation}
where $f_\mathrm{int}$ denotes the bilinear interpolation, which produces the Gaussian Texture with doubled resolution.
Thanks to the accurate surface tracking over the vertex level and texture level, the light-weight Gaussian texture super-resolution module can focus on the local regions and produces more intricate details such as yarn patterns which can be viewed in a much higher zoomin-level while incurring minimal overhead.
\begin{table}[t]
\renewcommand\tabcolsep{0.8pt}
\small
    \centering
    \caption{
        \textbf{Training status for each component}. We provide the status for each component and each loss function in the different training stages.
        The status of each component \good, indicating that the weights will be updated and the losses are used for training.
        Depth.A., Vert.A., Tex.A. and Tex.SR. denotes Depth Alignment (Sec.~\ref{subsec:surface}), Vertex Alignment (Sec.~\ref{subsec:triangle}), Texel Alignment (Sec.~\ref{subsec:texel}) and Texel Super Resolution (Sec.~\ref{subsec:srtexel}), respectively.
    }
    \vspace{-10pt}
    \label{tab:status}
    \begin{tabular}{|l|c|c|c|c|c|c|c|}
    \hline
    & \multicolumn{3}{c|}{Modules} & \multicolumn{4}{c|}{Supervision}  \\
    \cline{1-7}
    \hline
     & $\mathcal{F_\mathrm{eg}},\mathcal{F_\mathrm{delta}}$ & $\mathcal{E}_{\mathrm{app}},\mathcal{E}_{\mathrm{geo}}$ & $\mathcal{E}_{\mathrm{sr}}$ & $\mathcal{L}_\mathrm{temp}$ & $\mathcal{L}_\mathrm{gau}$ & $\mathcal{L}_\mathrm{cor-vrt}$ & $\mathcal{L}_\mathrm{cor-tex}$ \\
    \hline
    Depth.A. (Sec.~\ref{subsec:surface})  &  \good &   &       & \good  &        &          &      \\
    Vert.A. (Sec.~\ref{subsec:triangle})   &  \good &    &      & \good  &        &  \good   &      \\
    Tex.A. (Sec.~\ref{subsec:texel})    &        & \good &   &        & \good  &          & \good \\
    Tex.SR. (Sec.~\ref{subsec:srtexel})    &        & &\good    &        & \good  &          &       \\
    \hline
    \end{tabular}
\end{table}

%
%
%
%
\section{Results} \label{sec:results}
\par \noindent \textbf{Implementation Details.}
Our approach is implemented in PyTorch~\cite{paszke2017automatic}. 
For rendering, we adopt Analytical Splatting~\cite{liang2024analyticsplatting} for rasterizing the 3D Gaussian splats and employ Nvdiffrast~\cite{Laine2020diffrast} to render the textured meshes.
All the stages, competing methods and ablation alternatives are trained and tested on a server with two NVIDIA H100 graphics cards and a AMD EPYC 9554 CPU.
The depth alignment stage is trained for $360{,}000$ iterations using the Adam optimizer~\cite{kingma2017adam} with a learning rate of $1e^{-4}$ scheduled by a cosine decay, which takes around 12 hours.
The vertex alignment stage is trained for $360{,}000$ iterations using Adam optimizer with a learning rate of $5e^{-4}$ scheduled with a cosine decay scheduler, which takes around 12 hours.
The resolution of the Gaussian textures is set to $768\times 768$ for all the subjects, resulting in roughly $250k$ Gaussian splats in total.
After the Gaussian super resolution module, the Gaussian textures are upsampled to $1536 \times 1536$, i.e., approximately $1$ million Gaussian Splats in total.
To train the animatable Gaussian texture with texel-level alignment, following the open-sourced implementation in \citet{Pang_2024_CVPR}, it includes $15{,}000$ iterations of initialization before the main training, while the main training lasts for $2{,}000{,}000$ iterations with a learning rate of $1e^{-4}$.
The texel super resolution stage takes $1{,}000{,}000$ iterations with a learning rate of $1e^{-4}$.
The model is trained at an image resolution of $1620 \times 3072$ on random crops of size $810 \times 1536$.
\par \noindent \textbf{Dataset.} Our new dataset features five subjects wearing a diverse range of apparel, including loose-fitting tops and skirts.
Notably, unlike previous datasets that often include garments with limited texture complexity or solid colors, the subjects in our dataset wear clothing with rich and intricate patterns.
For each subject, we captured separate training and testing sequences in which they perform a variety of everyday motions, including jumping jacks, dancing, and boxing.
The sequences are captured using a multi-camera system consisting of $40$ synchronized and calibrated cameras, each recording at a resolution of $3240 \times 6144$ and a frame rate of $25$ fps.
The training sequences comprise approximately $17{,}000$ frames, while the testing sequences contain around $7{,}000$ frames.
Each frame of the captured videos is annotated with skeletal poses obtained using commercial 3D pose estimation software~\cite{captury}, foreground segmentation generated by Sapiens~\cite{khirodkar2024sapiens}, and pseudo ground-truth geometry reconstructed with NeuS2~\cite{neus2}.
Moreover, to facilitate comparison with existing methods, we additionally provide SMPL-X~\cite{SMPL-X:2019} parameters for the video frames.

\par \noindent \textbf{Metrics.} We adopt the Peak Signal-to-Noise Ratio (PSNR) metric to measure the quality of the rendered image.
Besides, we adopt the Structural Similarity Index (SSIM) and learned perceptual image patch similarity (LPIPS) ~\cite{zhang2018perceptual} that better mirrors human perception. 
Note that the metrics are computed at a resolution of $1620 \times 3072$, averaged over every 10th frame in the testing sequences, using two camera views that were excluded during training.
To assess the geometry reconstruction accuracy, we compute the Chamfer distance between generated mesh vertices (or 3D Gaussian point clouds) and the pseudo ground-truth reconstructions.

\subsection{Qualitative Results}
\begin{figure*}[h]
    \centering
    \includegraphics[width=\linewidth]{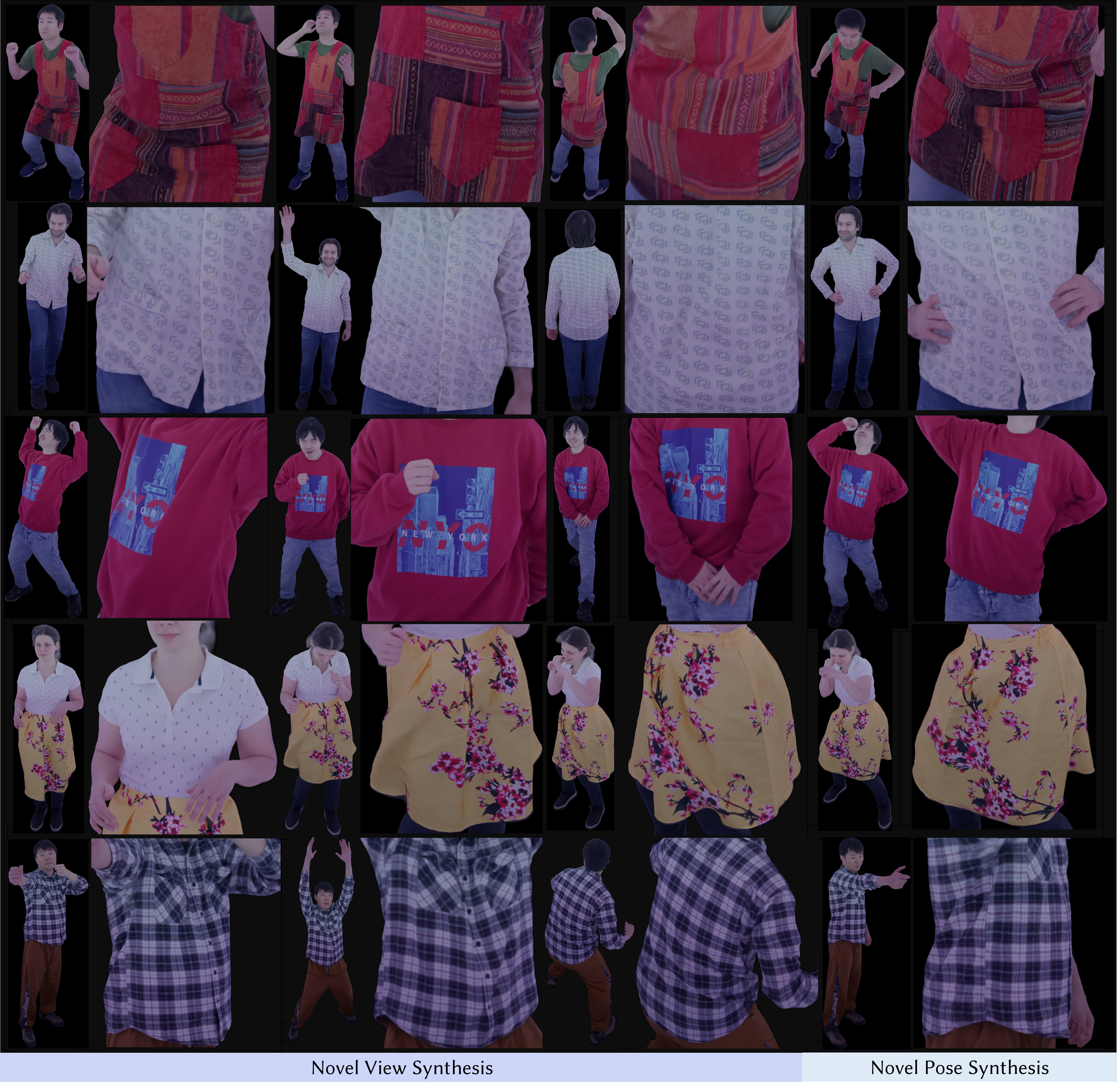}
    \vspace{-2em}
    \caption{
    %
    \textbf{Qualitative Rendering Results.}
    \titleabr performs well on both novel view and novel pose synthesis tasks, and manages to capture ultra details on human avatars, i.e texture patterns, cloth wrinkles. Please \textbf{zoom-in} to better observe the details.
    }
    \label{fig:qualitative}
    \vspace{-5pt}
\end{figure*}

%
%
\textbf{Image Synthesis.} Fig.~\ref{fig:qualitative} presents the image synthesis results. 
For both tight and loose fitting outfits, \titleabr faithfully recovers the garment wrinkles finest texture patterns in rendering for both novel views and also under novel poses.
\par
\begin{figure*}[h]
    \centering
    \includegraphics[width=\linewidth]{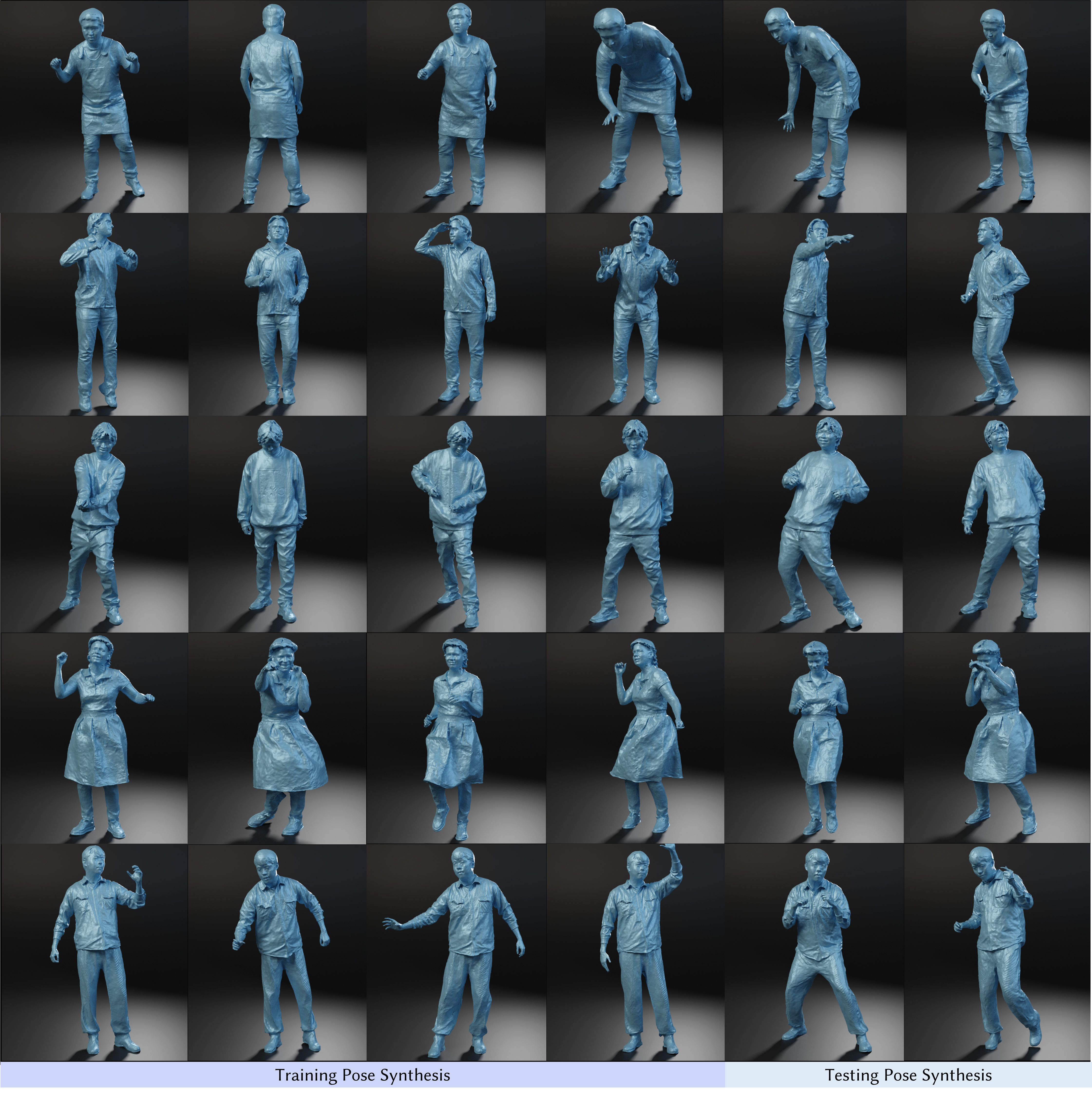}
    \vspace{-2em}
    \caption{
    \textbf{Qualitative Geometry Results.} For both training motions and testing motions unseen during taring, \titleabr generates clothing with realistic dynamics and vivid  detailed deformations. 
    Please \textbf{zoom-in} to better observe the details.
    }
    \label{fig:qualgeo}
\end{figure*}

\noindent\textbf{Geometry Synthesis.} Additionally, we show the geometry synthesis results in Fig.~\ref{fig:qualgeo}.
Note that \titleabr perseveres the clothing dynamics and the motion-aware detailed deformation of the clothing for the training poses.
For the novel motion unseen during training, \titleabr could produce plausible and vivid clothing dynamics, which is more prominent for loose clothing such as dresses.

%
%
%
%
%
%

%
%
\subsection{Comparisons}
%
\begin{table*}[t]
\small
    \centering
    \caption{\textbf{Quantitative Comparison}. 
        Here, we quantitatively compare our method with prior works on rendering quality and geometry accuracy on the training split and testing split across all sequences.
        We highlight the \ffst{best}, \fsnd{second-best}, and \ftrd{third-best} scores.
        \titleabr consistently outperforms previous methods in all metrics, especially for the perceptual matrix which better reflect the reconstruction of fine details.
    }
    \vspace{-3pt}
    \begin{tabular}{|l|c|c|c|c|c|c|c|c|}
    \hline
    &\multicolumn{4}{c|}{Training Pose}&\multicolumn{4}{c|}{Testing Pose}  \\
    \cline{2-9}
     \multirow{-2}{*}{Methods} & \textbf{PSNR} $\uparrow$ & \textbf{SSIM} $\uparrow$ & \textbf{LPIPS} $\downarrow$ & \textbf{Cham} $\downarrow$ & \textbf{PSNR} $\uparrow$  & \textbf{SSIM} $\uparrow$ & \textbf{LPIPS} $\downarrow$ & \textbf{Cham} $\downarrow$ \\ 
    \hline
     DDC.~\cite{habermann2021deeper}      & \trd{30.54}   & \trd{0.9215}     & 112.3         & \trd{1.579}   & 26.94        & \trd{0.8965}   & 127.6        & \trd{1.817}  \\
      MeshAvatar~\cite{chen2024meshavatar}       & 27.23         & 0.8873           & 105.7         & 1.898         & 25.98        & 0.8805         & 117.1        & 2.393  \\
      \hline
     TriHuman~\cite{zhu2023trihuman}      & 30.24         & 0.9166           & 83.04         & \snd{0.983}   & \trd{27.37}  & \fst{0.8977}   & \trd{106.3}  & \fst{1.495}  \\
     \hline
     ASH~\cite{Pang_2024_CVPR}       & \snd{35.96}   & \snd{0.9569}     & \snd{63.84}   & 2.100         & \snd{27.50}  & \snd{0.8974}   & 112.4        & 2.377  \\
     Animatable Gaussians~\cite{li2024animatable}        & 29.07         & 0.9042           & \trd{75.42}   & 2.323         & 26.06        & 0.8839         & \snd{103.3}  & 2.923  \\
     GaussianAvatar~\cite{hu2023gaussianavatar}    & 25.88         & 0.8884           & 127.2         & 3.273         & 25.26        & 0.8845         & 134.9        & 3.630 \\
     3DGS-Avatar~\cite{qian20233dgs}    & 25.55         & 0.8865           & 141.6         & 3.876         & 24.87        & 0.8822         & 146.7        & 4.129  \\
    \hline
    \textbf{Ours wo SR} & \fst{36.80}  & \fst{0.9657}  & \fst{41.90}  & \fst{0.876}     & \fst{27.66}  & 0.8943         & \fst{90.21}  & \snd{1.523} \\
    \textbf{Ours} & \fst{37.15}  & \fst{0.9681}  & \fst{35.02}  & \fst{0.876}{}     & \fst{27.68}  & 0.8937         & \fst{84.12}  & \snd{1.523} \\
    \hline
    \end{tabular}
    \label{tab:quantitive}
\end{table*}

%
\noindent\textbf{Competing Methods.} We conducted extensive benchmarking with various existing approaches on animatable avatars with different underlying shape representations.  
%
%
\par \noindent \textit{Mesh-based Approaches.} 
MeshAvatar~\cite{chen2024meshavatar} models the pose-dependent surface deformations and materials conditioned from orthogonal projected position maps.
DDC~\cite{habermann2021deeper} models motion-aware clothing dynamics using learned deformation and pose-conditioned texture maps.
%
%
\par \noindent \textit{Implicit-based Approaches.} TriHuman~\cite{zhu2023trihuman} is an implicit-based method that models dynamic geometry and appearance of the dynamic clothed human via a signed distance field and a color field, both conditioned on motion-dependent tri-planes in tangent space. 
It employs unbiased volume rendering~\cite{wang2021neus} to couple geometry and appearance fields.
%
%
\par \noindent \textit{Point-based Approaches.} 
ASH~\cite{Pang_2024_CVPR}, 3DGS-Avatar~\cite{qian20233dgs}, and GaussianAvatar~\cite{hu2023gaussianavatar} model detailed appearance using pose-dependent 3D Gaussians Splats defined in the texel space.
3DGS-Avatar~\cite{qian20233dgs} and GaussianAvatar~\cite{hu2023gaussianavatar} employ a parametric human body model~\cite{SMPL-X:2019} to represent coarse geometry, while ASH~\cite{Pang_2024_CVPR} leverages learned embedded deformations of the person-specific template mesh inspired by DDC~\cite{habermann2021deeper}.
Notably, we set the resolution of the Gaussian textures of ASH to $768\times 768$, which is the same for the texture resolution adopted in our work for fair comparison.
Animatable Gaussians~\cite{li2024animatable} models coarse-level geometry using a person-specific template mesh and represents appearance using 3D Gaussian splats inferred from a front-and-back orthogonally projected position map.
%
\begin{table*}[t]
\small
    \centering
    \caption{\textbf{Ablation Study}. 
        We compare our full method with design alternatives.
        By addressing depth misalignment as well as surface drift, and we observe consistent improvements in view synthesis and geometry generation accuracy across both the training and testing splits of the dataset. 
    }
    \vspace{-3pt}
    \begin{tabular}{|l|c|c|c|c|c|c|c|c|c|}
    \hline
    &\multicolumn{4}{c|}{Training Pose}&\multicolumn{4}{c|}{Testing Pose} \\
    \cline{2-9}
     \multirow{-2}{*}{Methods}& \textbf{PSNR} $\uparrow$ & \textbf{SSIM} $\uparrow$ & \textbf{LPIPS} $\downarrow$ & \textbf{Cham} $\downarrow$ & \textbf{PSNR} $\uparrow$  & \textbf{SSIM} $\uparrow$ & \textbf{LPIPS} $\downarrow$ & \textbf{Cham} $\downarrow$ \\ 
    \hline
     raw                       & 36.53        & 0.9516       & 52.32       & 1.058        & \trd{29.49} & \snd{0.9146}  & 85.29        & 1.344 \\
     raw + lat                 & 36.60        & 0.9548       & 50.22       & 0.995        & \snd{29.62} & \fst{0.9147}  & 81.74        & 1.300 \\
     raw + lat + longt.        & 34.60        & 0.9400       & 84.88       & 1.053        & 28.45       & \trd{0.9131}  & 121.6        & 1.418 \\
     raw + lat + rdr.          & 37.09        & 0.9601       & 48.52       & 0.973        & 29.36       & 0.9111        & 85.98        & 1.275 \\
     raw + lat + trk. 1st      & \trd{37.44}  & \trd{0.9631} & \trd{40.39} & \trd{0.955}  & 29.43       & 0.9094        & \trd{77.82}  & \trd{1.273} \\
     raw + lat + trk. 2nd      & \snd{38.15}  & \snd{0.9682} & \snd{29.74} & \snd{0.931}  & 29.38       & 0.9065        & \snd{71.07}  & \snd{1.258} \\
     \hline
    \textbf{Ours wo SR} & \fst{38.38}  & \fst{0.9697} & \fst{25.87} & \fst{0.771}  & \fst{29.74} & 0.9075 & \fst{63.62}  & \fst{1.151}  \\
    \textbf{Ours} & \fst{38.56}  & \fst{0.9704} & \fst{25.42} & \fst{0.771}  & \fst{29.74} & 0.9076 & \fst{62.71}  & \fst{1.151}   \\
    \hline
    \end{tabular}
    \label{tab:ablation}
    \vspace{-5pt}
\end{table*}

%
%
\par \noindent\textbf{Quantitative Comparison.} Tab.~\ref{tab:quantitive} presents the quantitative comparison against competing methods for novel view and novel pose synthesis.
For novel view synthesis, our method consistently outperforms competing methods in all metrics, with a particularly significant improvement in LPIPS, which better reflects human perception.
For novel pose synthesis, \titleabr achieves the highest PSNR and LPIPS scores, highlighting its robustness to unseen poses.
For geometry synthesis, \titleabr attains the lowest Chamfer distance on the training split and ranks second on the testing split with only a marginal difference to TriHuman. 
Thanks to our multi-level alignments, \titleabr achieves significant improvements over all Gaussian-based methods. 
%
%
\par
\begin{figure*}[t]
    \centering
    \includegraphics[width=\linewidth]{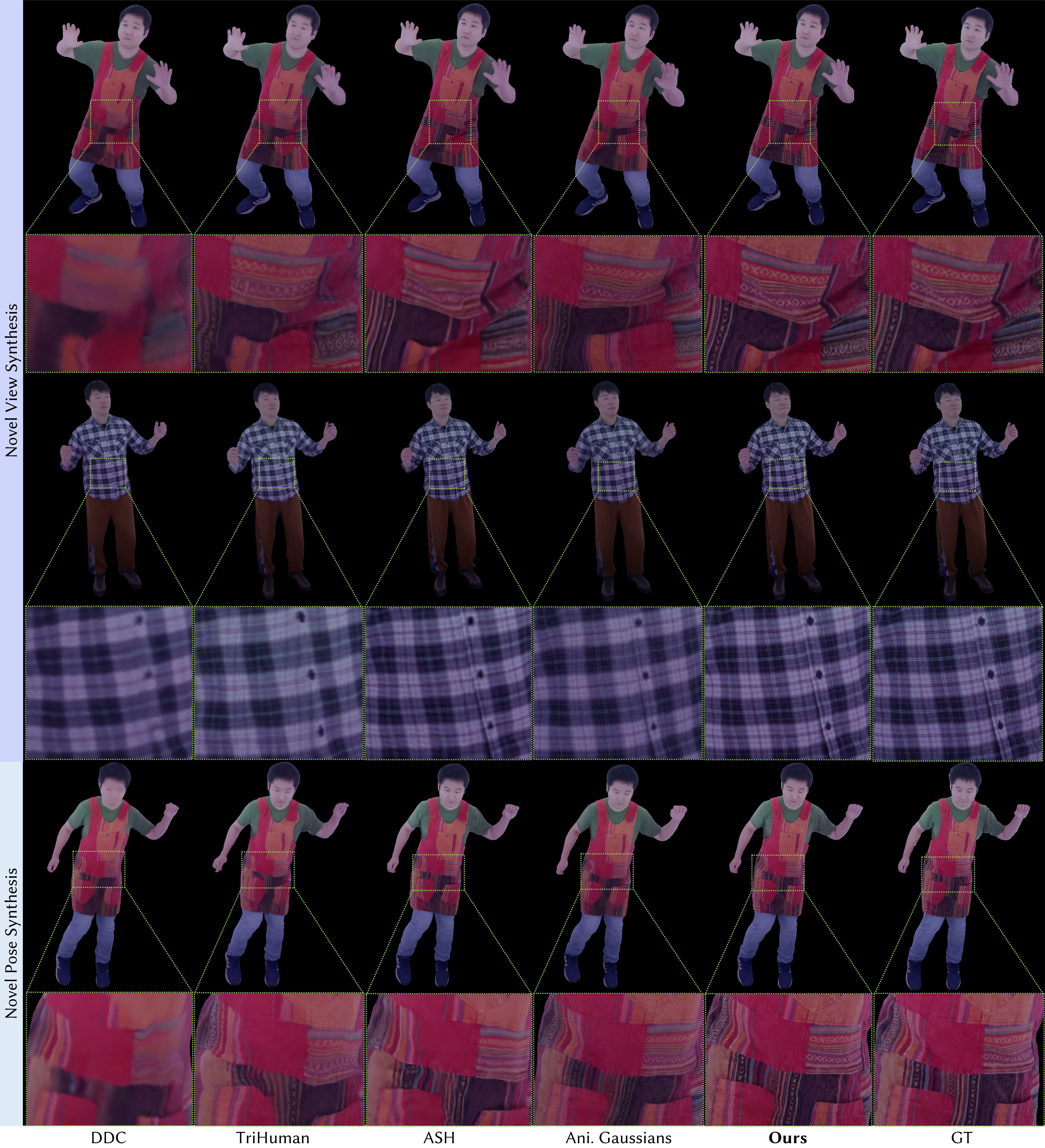}
    \vspace{-2.0em}
    \caption{
    \textbf{Qualitative Rendering Comparison.} We compare our approach with the competing approaches on novel view synthesis and novel pose generation.
    Compared with other methods, our approach preservers the best levels of details.
    Please \textbf{zoom-in} to better observe the details. 
    We refer to the supplemental document and video for additional qualitative comparisons with more methods and for the dynamic results.
    }
    \label{fig:comparison}
\end{figure*}

 \noindent\textbf{Qualitative Comparison.} Fig.~\ref{fig:comparison} presents the qualitative comparison on novel view and pose renderings.
DDC~\cite{habermann2021} and TriHuman~\cite{zhu2023trihuman} can not recover sharp texture patterns limited by their appearance representations. 
ASH~\cite{Pang_2024_CVPR} and Animatable Gaussians~\cite{li2024animatable} could recover high frequency details. However, due to the depth misalignment and surface drift, they tend to produce blurry and wrong texture patterns.
In stark contrast, \titleabr is able to reproduce more accurate and sharper texture patterns and cloth wrinkles.
%
%
%
%
%
%
%
\begin{figure*}[t]
    \centering
    \includegraphics[width=\linewidth]{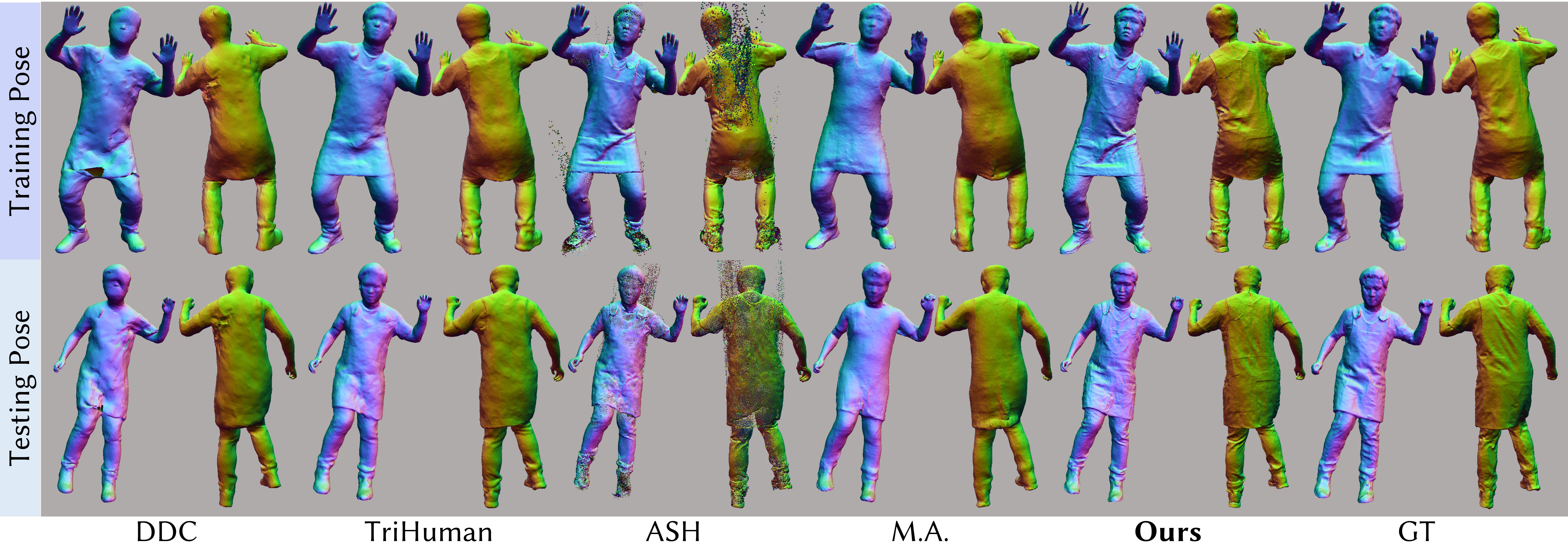}
    \vspace{-2em}
    \caption{
    \textbf{Qualitative Geometry Comparison.} We compare our approach with the competing approaches on novel view synthesis and novel pose generation.
    Compared with other methods, our approach preservers the best level of detail.
    Please \textbf{zoom-in} to better observe the details.
    }
    \label{fig:comparisongeo}
\end{figure*}
%
%
%
%
\subsection{Ablations}
\label{subsec:ablations}
To assess the impact of our core design choices, accounting for depth, vertex, and texel alignment, as well as the texel super-resolution, we conduct ablation studies by progressively building upon a baseline model with major components.
\par
\begin{figure}[h]
    \centering
    \includegraphics[width=\linewidth]{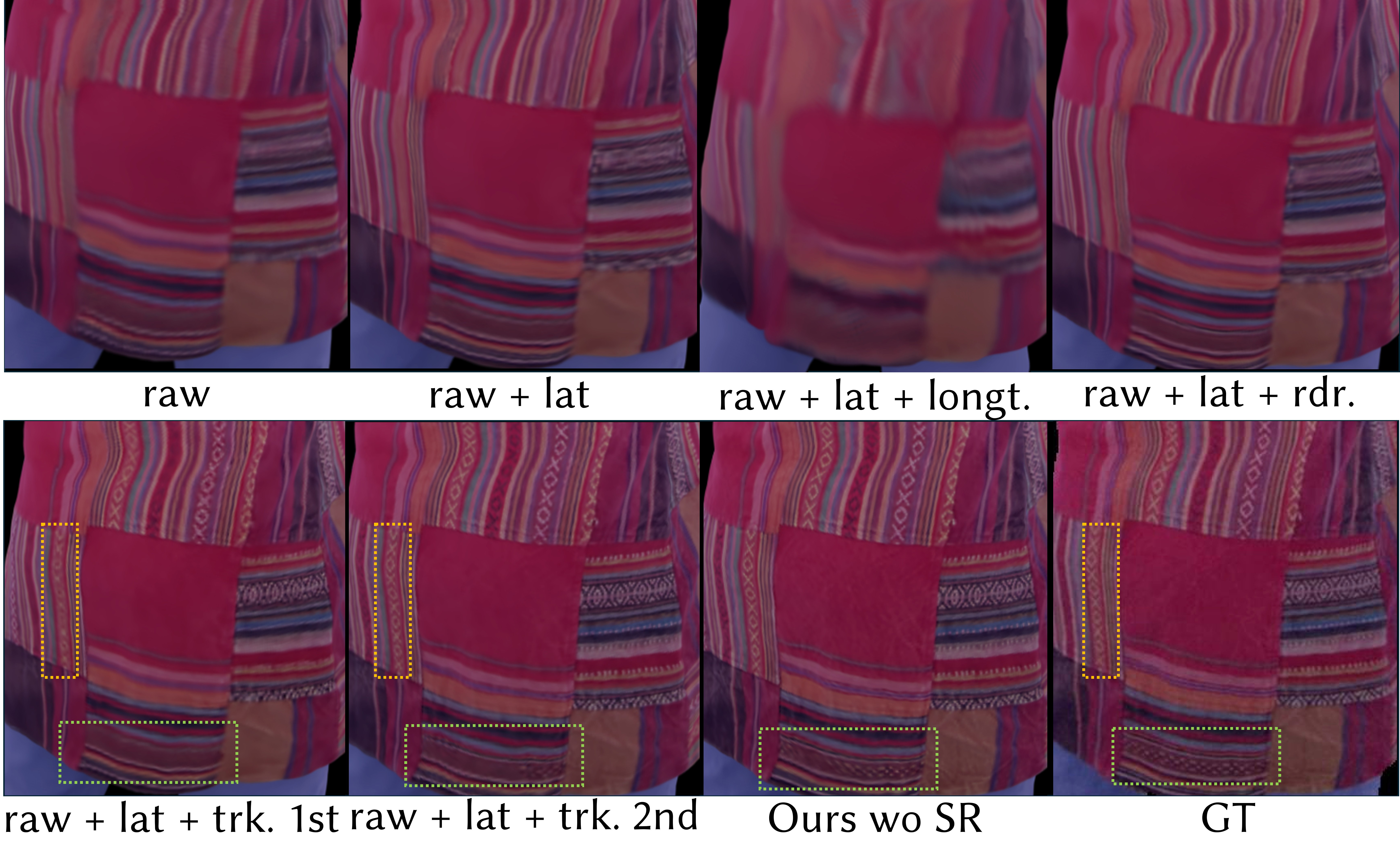}
    \vspace{-2em}
    \caption{
    \textbf{Qualitative Ablation on Renderings.}
    Depth alignment (\textbf{raw + lat}) enhances the fidelity of wrinkles and texture patterns.
    Compared with other alternatives (\textbf{raw + lat + longt.} and \textbf{raw + lat + rdr.}), our approach with vertex-level alignment (\textbf{raw + lat + trk. 1st} and \textbf{raw + lat + trk. 2nd}) further improves rendering quality, see the crossing highlighted in the yellow box.
    Finally, our model (\textbf{Ours wo SR}) with texel-level alignment recovers even the finest appearance details, such as the dots shown in the green box.
    Please \textbf{zoom-in} to better observe the details.
    }
    \label{fig:ablation}
\end{figure}

 \noindent \textbf{Baseline.} We begin with the baseline model (\textbf{raw}), namely, the animatable Gaussian textures introduced in Sec.~\ref{sec:preliminaries}.
Note that we adopt the Analytical Splatting~\cite{liang2024analyticsplatting} for rendering the Gaussaian textures for all the ablative experiments, which has proven capability on rendering at different scales.
As shown in Fig.~\ref{fig:ablation}, though the baseline model captures pose-dependent clothing wrinkles, it fails to reproduce fine texture patterns on the dress as the depth misalignment caused by the one-to-many mapping between the skeletal motion and surface deformations.
\par \noindent \textbf{Depth Alignment.} To address the depth misalignment, as mentioned in Sec.~\ref{subsec:surface}, we applied a per-frame latent $\boldsymbol{z}_f$ on top of the baseline approach as the additional input apart from the skeletal motion $\boldsymbol{\bar{\theta}}_{f}$ for predicting the drivable template mesh, termed as \textbf{raw + lat}. 
As shown in Tab.~\ref{tab:ablation}, this improvement in depth alignment effectively enhances the accuracy of surface reconstruction. More specifically, by applying the latent conditioning, the one-to-many ambiguity for the coarse template geometry, which attributes to the large scale dynamics error, is addressed. The improved template geometry also attributes to the improved performance on, both, novel view and novel pose synthesis.
However, as illustrated in Fig.~\ref{fig:ablation}, the absence of explicit surface correspondence supervision leads to on-surface drift, causing fine texture patterns to be missing in both novel-view and novel-pose renderings. 

%
%
\begin{figure}[h]
    \centering
    \includegraphics[width=0.98\linewidth]{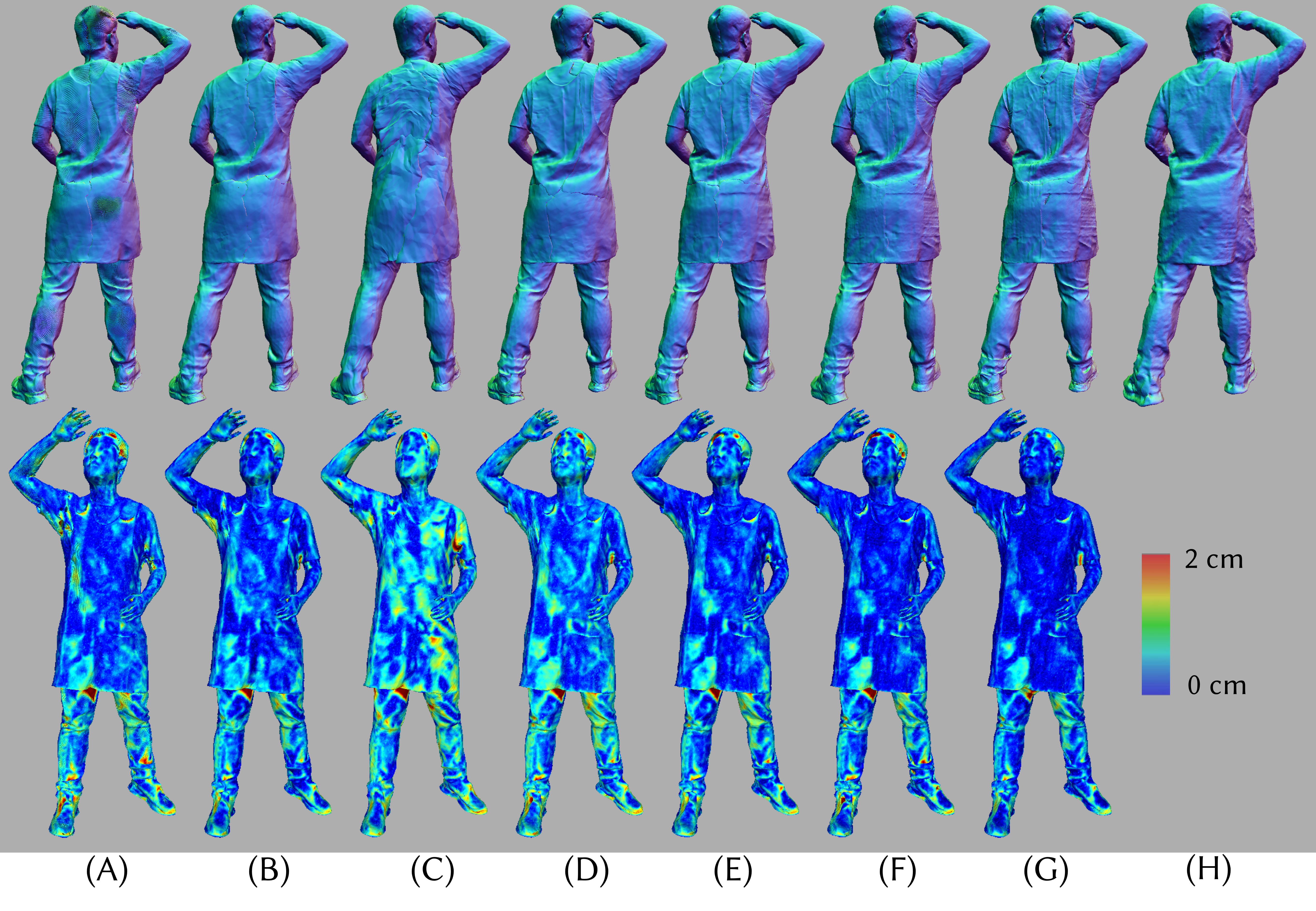}
    \vspace{-1em}
    \caption{
    \textbf{Qualitative Ablation on Geometry.}
    Naive Gaussian textures (\textbf{raw}) (A) exhibits highest error due to depth misalignment.
    (B) Introducing learnable latent(\textbf{raw + lat}) improves surface reconstruction accuracy.
    Comparing to naive correspondence tracking (\textbf{raw + lat + longt.}) (C) and differentiable rendering (\textbf{raw + lat + rdr.}) (D), our vertex-level alignment (\textbf{trk. 1st/2nd}) (E, F) achieves much higher accuracy.
    Our model (\textbf{Ours wo SR}) with texel-level alignment (G) achieves the highest reconstruction fidelity.
    }
    \label{fig:ablationgeo}
\end{figure}
%
%

%
%
%
%

%
\par \noindent \textbf{Vertex-level Alignment.} We compare two alternatives for supervising surface correspondence of the template mesh: using differentiable rendering losses (\textbf{raw + lat + rdr.}) and using correspondences directly extracted from the video point tracker~\cite{karaev2024cotracker} across the entire sequence (\textbf{raw + lat + logt.}).

To supervise surface correspondence with a differentiable rendering loss (\textbf{raw + lat + rdr.}), we augment the drivable template loss $\mathcal{L}_\mathrm{temp}$ 
by rendering the static textures $\mathbf{T}_{\mathrm{0}}$ with Nvdiffrast~\cite{Laine2020diffrast} and applying a $L1$ photometric loss between the rendered static-textured template mesh and the ground truth image.
As shown in Tab.~\ref{tab:ablation} and Fig.~\ref{fig:ablation}, applying differentiable rendering (\textbf{raw + lat + rdr.}) yields only limited quantitative and qualitative improvements, as the optimization often gets trapped in local minima caused by diverse shading and surface deformations.

As another design alternative, we employ tracked correspondences (\textbf{raw + lat + logt.}) obtained directly from a video point tracker from different views as supervision.
Specifically, the vertices of the posed template mesh in the first frame are projected into all training camera views and serve as the initialization for point tracking in each view.
Then, we adopt the video point tracker to track the point movements starting from the first frame to the end of the sequence.
The tracked points/vertices in the image space are unprojected to 3D to served as the supervision for the template mesh vertices.
As is shown in Fig.~\ref{fig:ablation} due to the accumulated errors, the naively tracked correspondence could not provide meaningful supervision, and leads to even more blurry results.

In stark contrast, our approach leverages the animatable template mesh to aggregate and refine correspondences (\textbf{raw + lat + trk. 1st}) across multiple views and frames, yielding more accurate correspondence supervision for the template vertices.
This more accurate and temporally consistent correspondence supervision is reflected in the recovery of fine texture details that would otherwise appear blurred without proper correspondence supervision, as shown in Fig.~\ref{fig:ablation}.
Moreover, as shown in Tab.~\ref{tab:ablation}, the quantitative performance is significantly improved for both novel-view and novel-pose rendering tasks, with especially large gains in the LPIPS metric, which better reflects human perception.

We further iterate the vertex alignment for a second time (\textbf{raw + lat + trk. 2st}), which further improves both qualitative and quantitative performance.

\par \noindent \textbf{Texel-level Alignment.} As discussed in Sec.~\ref{subsec:texel}, texel-level alignment provides dense and precise supervision by establishing detailed correspondences between texels and the ground-truth surface.
By leveraging the texel-level correspondences, our approach (\textbf{Ours wo SR}) achieves a substantial improvement in surface reconstruction accuracy, along with further gains in, both, accuracy and visual quality for novel view and pose synthesis.

\par
\begin{figure*}[h]
    \centering
    \includegraphics[width=\linewidth]{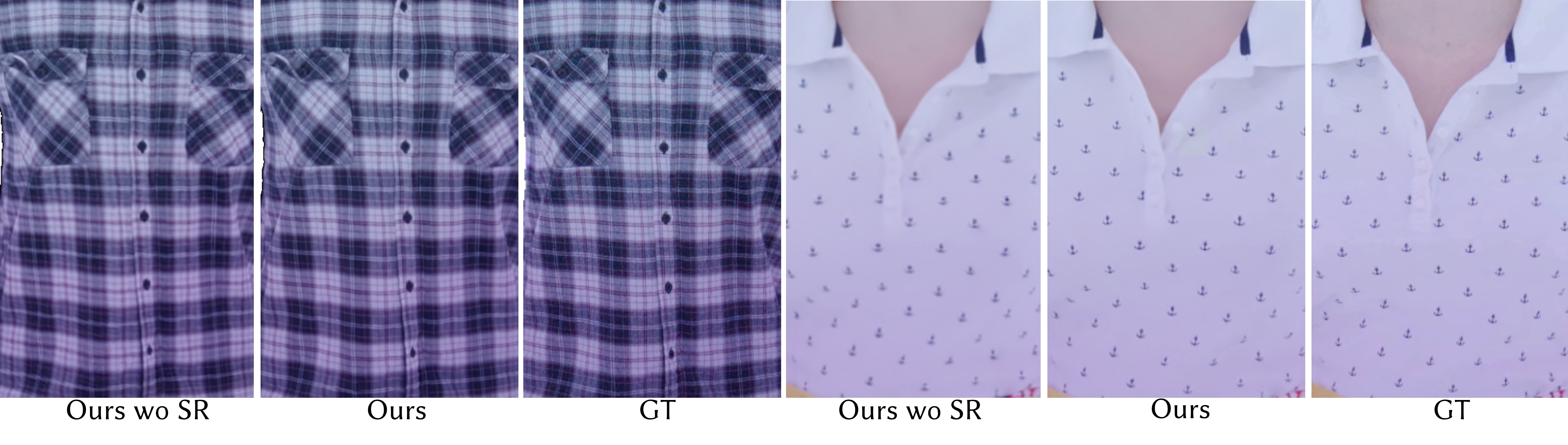}
    \vspace{-3em}
    \caption{
    \textbf{Qualitative ablation for texel super resolution.} By applying texel super-resolution, our model could further preserve the tiny patterns on the clothing. \textbf{Please \textbf{zoom-in} to better observe the details.} 
    }
    \label{fig:ablation_sr}
\end{figure*}

 \noindent \textbf{Texel Super Resolution.} Through surface-, vertex-, and texel-level alignment, our method achieves unprecedented quality in both rendering and geometry.
However, for challenging cases, details like fine texture patterns and yarn structures cannot be adequately captured with Gaussian textures with the original resolution as illustrated in Fig.~\ref{fig:ablation_sr}.
Therefore, as mentioned in Sec.~\ref{subsec:srtexel}, we apply the Gaussian Texture super-resolution module to the Gaussian Textures, denoted as \textbf{Ours}.
As can be seen from Fig.~\ref{fig:ablation_sr}, by applying the Gaussian Texture super-resolution module, the tiny glyphs on the clothing, could be properly reconstructed.
Moreover, it also leads to quantitative improvements on both novel view and novel pose synthesis task.
It is worth mentioning that, since the network only models the residual of the Gaussian Splatting parameters w.r.t. the original Gaussian Textures, and employs a lightweight architecture, its computational overhead remains minimal (\textbf{Ours wo SR} runs at $21$ fps, while \textbf{Ours} runs at $18$ fps).

%
%
%
%
\section{Applications}
\subsection{VR Telepresence}
\begin{figure}[t]
    \centering
    \includegraphics[width=\linewidth]{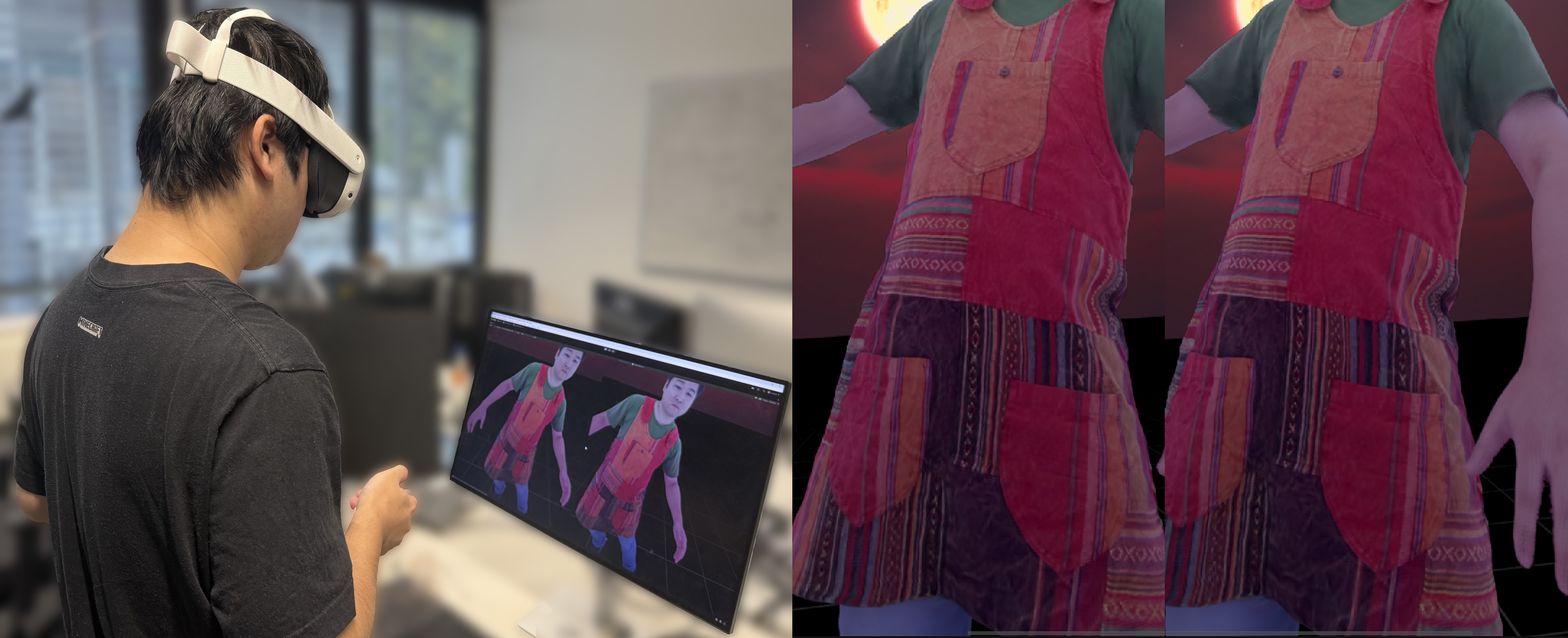}
    \vspace{-2em}
    \caption{
    \textbf{VR Application.} Users may inspect the dynamic avatar in VR headsets. The high-quality avatar geometry and appearance provides unprecedented immersive experiences, which may further boost the applications such as teleconference, remote assistance and so on.
    }
    \label{fig:appvr}
\end{figure}

We implemented a VR demo to visualize the results generated by \titleabr.
The system is built with Unity3D and integrates the gsplat Gaussian Splat~\cite{ye2025gsplat} renderer. 
For each frame, the Gaussian Splatting results are precomputed and stored on disk. 
The Unity3D backend receives the VR headset’s pose in real time and renders the corresponding view, which is then streamed from the workstation to the headset.
As shown in Fig.~\ref{fig:appvr}, users can walk around to inspect the virtual character.
%
%
%
%
%

\subsection{Motion Editing}
The animatable avatar generated by \titleabr paves the way for creating photorealistic, high-quality content, enabling users to animate the character with desired motions and render it from arbitrary viewpoints and zoom-in levels.
To facilitate animation creation, we present UMA-Viewer, a tool that enables users to inspect recorded sequences as well as edit and visualize renderings of newly generated motion sequences.
The front-end system is implemented with Viser~\cite{yi2025viser} and runs on a personal computer.
The backend is implemented in PyTorch, generating renderings on the fly based on the skeleton motion and camera parameters received from the front-end, and streaming results back at an interactive frame rate.

\noindent \textbf{Inspecting recorded sequences.} UMA-Viewer enables users to inspect characters driven by recorded motion sequences.
Specifically, users can playback entire sequences from various camera viewpoints.
For any given frame, the character can be freely rotated and zoomed in for detailed inspection.
Furthermore, the system supports playback along user-defined camera trajectories, with adjustable zoom-in levels along the camera path.

%
%
\begin{figure}[t]
    \centering
    \includegraphics[width=\linewidth]{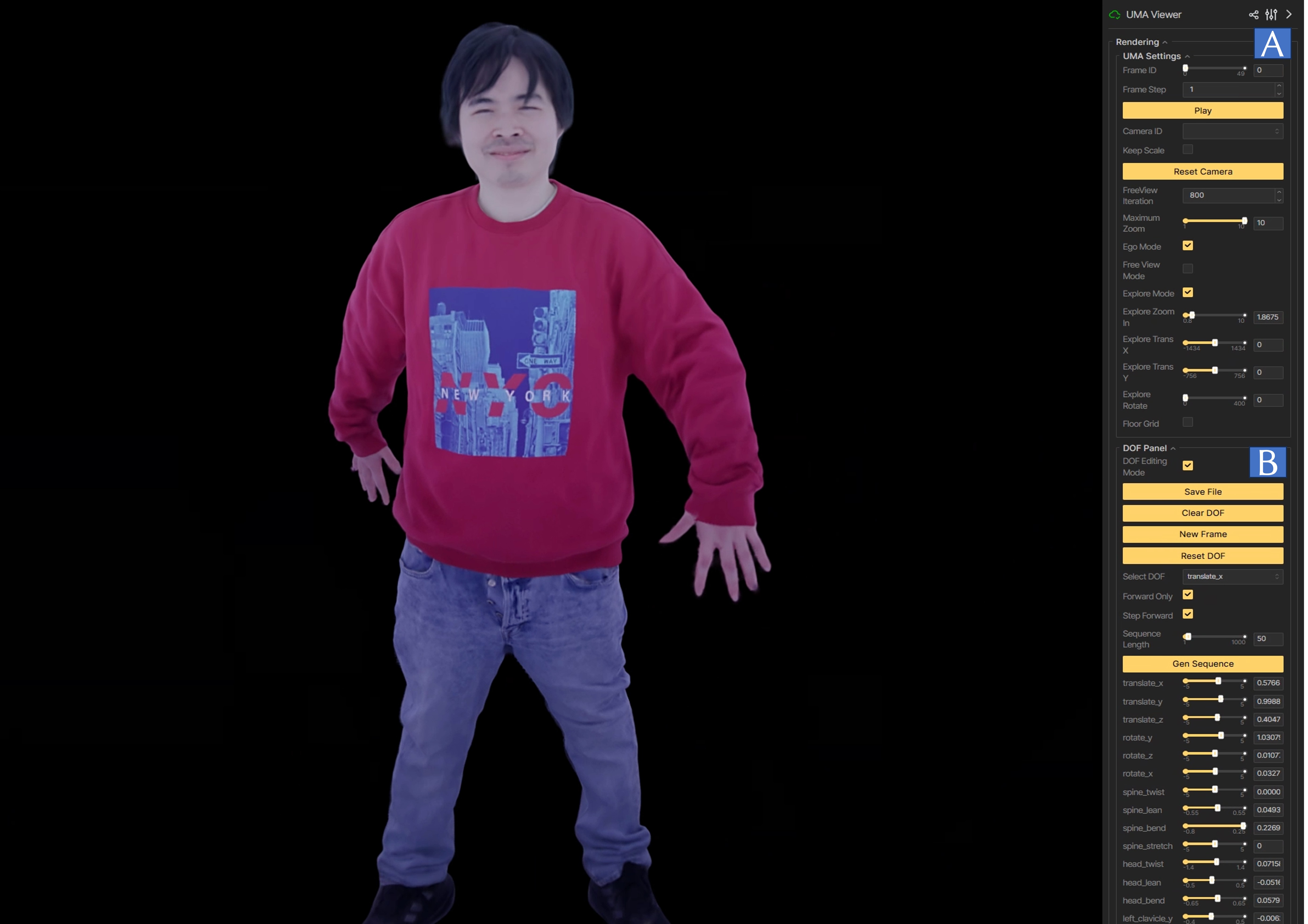}
    \vspace{-2em}
    \caption{
    \textbf{User interface for UMA-Viewer.} UMA-Viewer offers is an interactive system to inspect the recorded sequence and create novel motion sequences. (A) The control panel for changing the rendering settings. (B) The DOF panel editing and generating motion sequences. 
    }
    \label{fig:apppanel}
\end{figure}
%
%

\noindent \textbf{Creating Novel Sequences.} Beyond inspecting recorded motion sequences, UMA-Viewer also enables users to interactively create and examine novel motion sequences.
Specifically, users can assign values to each degree of freedom (DOF) for newly created motion frames.
Once the DOF is modified, the system will immediately visualize the character rendering.

\begin{figure}[t]
    \centering
    \includegraphics[width=\linewidth]{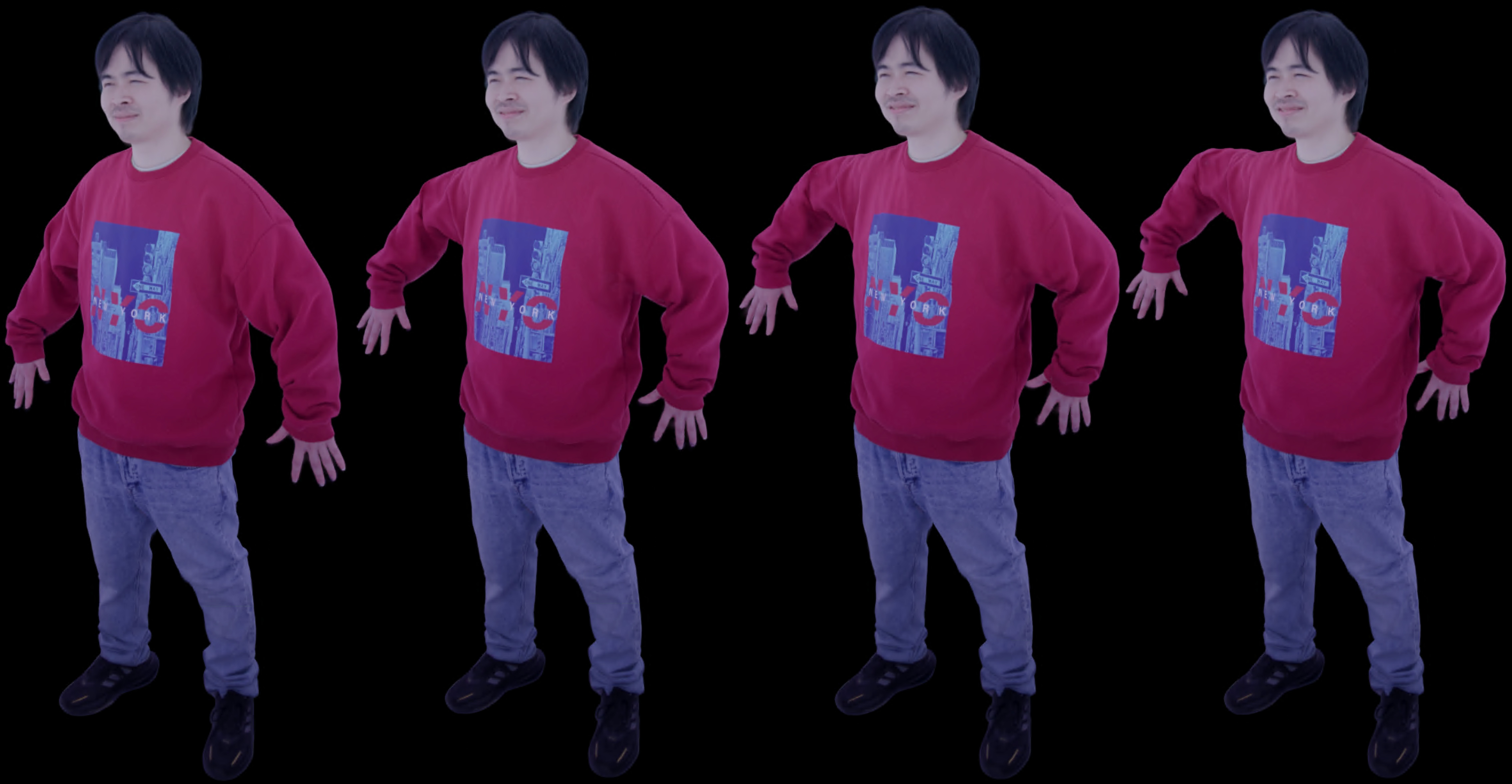}
    \vspace{-2em}
    \caption{
    \textbf{Motion Editing Results.} The rendering with motion generated with UMA-Viewer, note the clothing wrinkles caused by bending the elbow and raising the arm.
    }
    \label{fig:appmotion}
\end{figure}

Editing motion sequences frame by frame is a laborious and time-consuming task.
To address this, UMA-Viewer allows users to efficiently generate motion sequences through key-framing.
Specifically, for each DOF to be edited, users can specify the step size and the sequence length for motion sequence.
As illustrated in Fig.~\ref{fig:appmotion}, the resulting animatable character under the edited motion sequence exhibits plausible wrinkle dynamics and detailed textures.

%
%
%
%

\subsection{Motion Retargeting.} 
\begin{figure*}[h]
    \centering
    \includegraphics[width=0.98\linewidth]{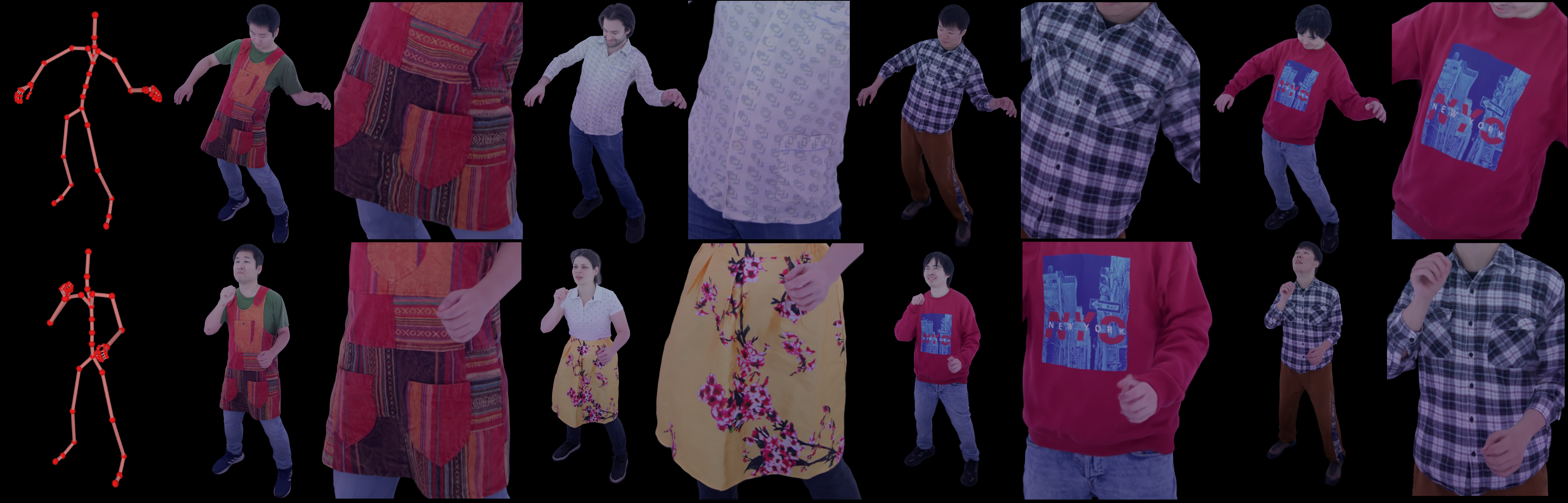}
    \vspace{-5pt}
    \caption{
    \textbf{Motion retargeting.} \titleabr enables animating different characters using the same skeletal motion depicted in the leftmost column, while faithfully preserving fine texture details and producing photorealistic wrinkles.
    }
    \label{fig:supplretarget}
\end{figure*}

\titleabr is trained on high-quality, long multi-view sequences encompassing a diverse range of motions, enabling it to generalize effectively to novel poses.
In Fig.~\ref{fig:supplretarget}, we present motion re-targeting results where all characters are driven by the same skeletal motion.
Despite the motions being unseen during training, our method successfully renders photorealistic avatars with intricate appearance details and realistic wrinkles.
%
%
%
%
\section{Limitations and Future Work}
Though \titleabr makes clear improvements towards ultra-detailed human avatar modeling, there are still several open challenges left to solve. 
Since the clothed human avatar is driven with a single-layered person specific template, it does not support changing of the outfits.
A layered representation which separates the modeling of garments and body could support the virtual try-on applications.
Moreover, since the surface deformation of the garments is only conditioned on the skeletal motions, \titleabr currently does not support physical effects due to the external forces such as the interaction between clothing and objects.
Besides, although \titleabr reconstructs high-fidelity surface geometry and apperance, which is crucial for applications for relighting. 
Later efforts could integrate the estimation of lights and materials to support the relighting of the ultra-high resolution rendering of clothed humans.
Lastly, \titleabr primarily allows the user to control the skeletal motion, i.e., the body pose and hand poses.
Future work could explore more expressive control including for example facial expressions.

\section{Conclusion}
\label{sec:conclusion}
We presented \titleabr, a novel method for ultra-detailed animatable human avatar creation.
\titleabr achieves superior rendering quality over state-of-the-art human avatar approaches and faithfully captures the finest cloth wrinkles and texture patterns.
At the core, we analyzed the reason why existing approaches on the animatable characters does not recovers the finest appearances details, which is due to the tracking errors from multiple levels, i.e., surface, vertex and texel levels.
To this end, we introduced a multi-level alignment approach for better tracking quality of humans.
Notably, we firstly introduce the additional latent condition on the coarse geometry, which is used for modeling the stochastic affects that cannot be described solely by skeletal motions.
With the improved avatar representation, we further introduce adopting the foundational point tracker as the supervision for the surface deformation to replace the pixel-wise loss for multiple granite of geometry.
Though a multi-level geometry refinement, we progressively improves the tracking of the surface and therefore improves rendering performance, especially for the finer level details.
We believe \titleabr is a significant step towards ultra-realistic human avatar modeling supporting future innovations in VR/AR, film making, and tele-conferences.

\bibliographystyle{ACM-Reference-Format}
\bibliography{main}


\begin{thebibliography}{78}


\ifx \showCODEN    \undefined \def \showCODEN     #1{\unskip}     \fi
\ifx \showISBNx    \undefined \def \showISBNx     #1{\unskip}     \fi
\ifx \showISBNxiii \undefined \def \showISBNxiii  #1{\unskip}     \fi
\ifx \showISSN     \undefined \def \showISSN      #1{\unskip}     \fi
\ifx \showLCCN     \undefined \def \showLCCN      #1{\unskip}     \fi
\ifx \shownote     \undefined \def \shownote      #1{#1}          \fi
\ifx \showarticletitle \undefined \def \showarticletitle #1{#1}   \fi
\ifx \showURL      \undefined \def \showURL       {\relax}        \fi
\providecommand\bibfield[2]{#2}
\providecommand\bibinfo[2]{#2}
\providecommand\natexlab[1]{#1}
\providecommand\showeprint[2][]{arXiv:#2}

\bibitem[Alldieck et~al\mbox{.}(2019)]%
        {alldieck19}
\bibfield{author}{\bibinfo{person}{Thiemo Alldieck}, \bibinfo{person}{Marcus Magnor}, \bibinfo{person}{Bharat~Lal Bhatnagar}, \bibinfo{person}{Christian Theobalt}, {and} \bibinfo{person}{Gerard Pons-Moll}.} \bibinfo{year}{2019}\natexlab{}.
\newblock \showarticletitle{Learning to Reconstruct People in Clothing from a Single {RGB} Camera}. In \bibinfo{booktitle}{\emph{IEEE Conf. Comput. Vis. Pattern Recog.}} \bibinfo{pages}{1175--1186}.
\newblock


\bibitem[Alldieck et~al\mbox{.}(2018)]%
        {alldieck18b}
\bibfield{author}{\bibinfo{person}{Thiemo Alldieck}, \bibinfo{person}{Marcus Magnor}, \bibinfo{person}{Weipeng Xu}, \bibinfo{person}{Christian Theobalt}, {and} \bibinfo{person}{Gerard Pons-Moll}.} \bibinfo{year}{2018}\natexlab{}.
\newblock \showarticletitle{Detailed Human Avatars from Monocular Video}. In \bibinfo{booktitle}{\emph{International Conference on 3D Vision}}. \bibinfo{pages}{98--109}.
\newblock
\href{https://doi.org/10.1109/3{DV}.2018.00022}{doi:\nolinkurl{10.1109/3{DV}.2018.00022}}


\bibitem[Bagautdinov et~al\mbox{.}(2021)]%
        {bagautdinov2021driving}
\bibfield{author}{\bibinfo{person}{Timur Bagautdinov}, \bibinfo{person}{Chenglei Wu}, \bibinfo{person}{Tomas Simon}, \bibinfo{person}{Fabian Prada}, \bibinfo{person}{Takaaki Shiratori}, \bibinfo{person}{Shih-En Wei}, \bibinfo{person}{Weipeng Xu}, \bibinfo{person}{Yaser Sheikh}, {and} \bibinfo{person}{Jason Saragih}.} \bibinfo{year}{2021}\natexlab{}.
\newblock \showarticletitle{Driving-signal aware full-body avatars}.
\newblock \bibinfo{journal}{\emph{ACM Transactions on Graphics (TOG)}} \bibinfo{volume}{40}, \bibinfo{number}{4} (\bibinfo{year}{2021}), \bibinfo{pages}{1--17}.
\newblock


\bibitem[Casas et~al\mbox{.}(2014)]%
        {casas14}
\bibfield{author}{\bibinfo{person}{Dan Casas}, \bibinfo{person}{Marco Volino}, \bibinfo{person}{John Collomosse}, {and} \bibinfo{person}{Adrian Hilton}.} \bibinfo{year}{2014}\natexlab{}.
\newblock \showarticletitle{4D Video Textures for Interactive Character Appearance}.
\newblock \bibinfo{journal}{\emph{Comput. Graph. Forum}} \bibinfo{volume}{33}, \bibinfo{number}{2} (\bibinfo{date}{May} \bibinfo{year}{2014}), \bibinfo{pages}{371–380}.
\newblock
\showISSN{0167-7055}
\href{https://doi.org/10.1111/cgf.12296}{doi:\nolinkurl{10.1111/cgf.12296}}


\bibitem[Chen et~al\mbox{.}(2024)]%
        {chen2024meshavatar}
\bibfield{author}{\bibinfo{person}{Yushuo Chen}, \bibinfo{person}{Zerong Zheng}, \bibinfo{person}{Zhe Li}, \bibinfo{person}{Chao Xu}, {and} \bibinfo{person}{Yebin Liu}.} \bibinfo{year}{2024}\natexlab{}.
\newblock \showarticletitle{Meshavatar: Learning high-quality triangular human avatars from multi-view videos}. In \bibinfo{booktitle}{\emph{Eur. Conf. Comput. Vis.}} Springer, \bibinfo{pages}{250--269}.
\newblock


\bibitem[Gao et~al\mbox{.}(2023)]%
        {NNA}
\bibfield{author}{\bibinfo{person}{Qingzhe Gao}, \bibinfo{person}{Yiming Wang}, \bibinfo{person}{Libin Liu}, \bibinfo{person}{Lingjie Liu}, \bibinfo{person}{Christian Theobalt}, {and} \bibinfo{person}{Baoquan Chen}.} \bibinfo{year}{2023}\natexlab{}.
\newblock \showarticletitle{Neural novel actor: Learning a generalized animatable neural representation for human actors}.
\newblock \bibinfo{journal}{\emph{IEEE Trans. Vis. Comput. Graph.}} (\bibinfo{year}{2023}).
\newblock


\bibitem[Guan et~al\mbox{.}(2012)]%
        {guan2012drape}
\bibfield{author}{\bibinfo{person}{Peng Guan}, \bibinfo{person}{Loretta Reiss}, \bibinfo{person}{David~A Hirshberg}, \bibinfo{person}{Alexander Weiss}, {and} \bibinfo{person}{Michael~J Black}.} \bibinfo{year}{2012}\natexlab{}.
\newblock \showarticletitle{Drape: Dressing any person}.
\newblock \bibinfo{journal}{\emph{TOG}} \bibinfo{volume}{31}, \bibinfo{number}{4} (\bibinfo{year}{2012}), \bibinfo{pages}{1--10}.
\newblock


\bibitem[Habermann et~al\mbox{.}(2023)]%
        {habermann2023hdhumans}
\bibfield{author}{\bibinfo{person}{Marc Habermann}, \bibinfo{person}{Lingjie Liu}, \bibinfo{person}{Weipeng Xu}, \bibinfo{person}{Gerard Pons-Moll}, \bibinfo{person}{Michael Zollhoefer}, {and} \bibinfo{person}{Christian Theobalt}.} \bibinfo{year}{2023}\natexlab{}.
\newblock \showarticletitle{Hdhumans: A hybrid approach for high-fidelity digital humans}.
\newblock \bibinfo{journal}{\emph{Proceedings of the ACM on Computer Graphics and Interactive Techniques}} \bibinfo{volume}{6}, \bibinfo{number}{3} (\bibinfo{year}{2023}), \bibinfo{pages}{1--23}.
\newblock


\bibitem[Habermann et~al\mbox{.}(2021a)]%
        {habermann2021}
\bibfield{author}{\bibinfo{person}{Marc Habermann}, \bibinfo{person}{Lingjie Liu}, \bibinfo{person}{Weipeng Xu}, \bibinfo{person}{Michael Zollhoefer}, \bibinfo{person}{Gerard Pons-Moll}, {and} \bibinfo{person}{Christian Theobalt}.} \bibinfo{year}{2021}\natexlab{a}.
\newblock \showarticletitle{Real-time Deep Dynamic Characters}.
\newblock \bibinfo{journal}{\emph{ACM Trans. Graph.}} \bibinfo{volume}{40}, \bibinfo{number}{4}, Article \bibinfo{articleno}{94} (\bibinfo{date}{aug} \bibinfo{year}{2021}).
\newblock


\bibitem[Habermann et~al\mbox{.}(2019)]%
        {habermann2019livecap}
\bibfield{author}{\bibinfo{person}{Marc Habermann}, \bibinfo{person}{Weipeng Xu}, \bibinfo{person}{Michael Zollhoefer}, \bibinfo{person}{Gerard Pons-Moll}, {and} \bibinfo{person}{Christian Theobalt}.} \bibinfo{year}{2019}\natexlab{}.
\newblock \showarticletitle{Livecap: Real-time human performance capture from monocular video}.
\newblock \bibinfo{journal}{\emph{ACM Transactions On Graphics (TOG)}} \bibinfo{volume}{38}, \bibinfo{number}{2} (\bibinfo{year}{2019}), \bibinfo{pages}{1--17}.
\newblock


\bibitem[Habermann et~al\mbox{.}(2021b)]%
        {habermann2021deeper}
\bibfield{author}{\bibinfo{person}{Marc Habermann}, \bibinfo{person}{Weipeng Xu}, \bibinfo{person}{Michael Zollhoefer}, \bibinfo{person}{Gerard Pons-Moll}, {and} \bibinfo{person}{Christian Theobalt}.} \bibinfo{year}{2021}\natexlab{b}.
\newblock \showarticletitle{A deeper look into deepcap}.
\newblock \bibinfo{journal}{\emph{IEEE Transactions on Pattern Analysis and Machine Intelligence}} \bibinfo{volume}{45}, \bibinfo{number}{4} (\bibinfo{year}{2021}), \bibinfo{pages}{4009--4022}.
\newblock


\bibitem[Habermann et~al\mbox{.}(2020)]%
        {habermann2020deepcap}
\bibfield{author}{\bibinfo{person}{Marc Habermann}, \bibinfo{person}{Weipeng Xu}, \bibinfo{person}{Michael Zollhofer}, \bibinfo{person}{Gerard Pons-Moll}, {and} \bibinfo{person}{Christian Theobalt}.} \bibinfo{year}{2020}\natexlab{}.
\newblock \showarticletitle{Deepcap: Monocular human performance capture using weak supervision}. In \bibinfo{booktitle}{\emph{IEEE Conf. Comput. Vis. Pattern Recog.}} \bibinfo{pages}{5052--5063}.
\newblock


\bibitem[Hu et~al\mbox{.}(2024)]%
        {hu2023gaussianavatar}
\bibfield{author}{\bibinfo{person}{Liangxiao Hu}, \bibinfo{person}{Hongwen Zhang}, \bibinfo{person}{Yuxiang Zhang}, \bibinfo{person}{Boyao Zhou}, \bibinfo{person}{Boning Liu}, \bibinfo{person}{Shengping Zhang}, {and} \bibinfo{person}{Liqiang Nie}.} \bibinfo{year}{2024}\natexlab{}.
\newblock \showarticletitle{Gaussianavatar: Towards realistic human avatar modeling from a single video via animatable 3d gaussians}. In \bibinfo{booktitle}{\emph{CVPR}}.
\newblock


\bibitem[Hu and Liu(2024)]%
        {hu2023gauhuman}
\bibfield{author}{\bibinfo{person}{Shoukang Hu} {and} \bibinfo{person}{Ziwei Liu}.} \bibinfo{year}{2024}\natexlab{}.
\newblock \showarticletitle{Gauhuman: Articulated gaussian splatting from monocular human videos}. In \bibinfo{booktitle}{\emph{CVPR}}.
\newblock


\bibitem[I\c{s}{\i}k et~al\mbox{.}(2023)]%
        {isik2023humanrf}
\bibfield{author}{\bibinfo{person}{Mustafa I\c{s}{\i}k}, \bibinfo{person}{Martin Runz}, \bibinfo{person}{Markos Georgopoulos}, \bibinfo{person}{Taras Khakhulin}, \bibinfo{person}{Jonathan Starck}, \bibinfo{person}{Lourdes Agapito}, {and} \bibinfo{person}{Matthias Niessner}.} \bibinfo{year}{2023}\natexlab{}.
\newblock \showarticletitle{HumanRF: High-Fidelity Neural Radiance Fields for Humans in Motion}.
\newblock \bibinfo{journal}{\emph{ACM Trans. Graph.}} \bibinfo{volume}{42}, \bibinfo{number}{4} (\bibinfo{year}{2023}), \bibinfo{pages}{1--12}.
\newblock
\href{https://doi.org/10.1145/3592415}{doi:\nolinkurl{10.1145/3592415}}


\bibitem[Jiang et~al\mbox{.}(2025)]%
        {jiang2025reperformer}
\bibfield{author}{\bibinfo{person}{Yuheng Jiang}, \bibinfo{person}{Zhehao Shen}, \bibinfo{person}{Chengcheng Guo}, \bibinfo{person}{Yu Hong}, \bibinfo{person}{Zhuo Su}, \bibinfo{person}{Yingliang Zhang}, \bibinfo{person}{Marc Habermann}, {and} \bibinfo{person}{Lan Xu}.} \bibinfo{year}{2025}\natexlab{}.
\newblock \showarticletitle{RePerformer: Immersive Human-centric Volumetric Videos from Playback to Photoreal Reperformance}.
\newblock \bibinfo{journal}{\emph{arXiv preprint arXiv:2503.12242}} (\bibinfo{year}{2025}).
\newblock


\bibitem[Joo et~al\mbox{.}(2018)]%
        {TotalCapture2018}
\bibfield{author}{\bibinfo{person}{Hanbyul Joo}, \bibinfo{person}{Tomas Simon}, {and} \bibinfo{person}{Yaser Sheikh}.} \bibinfo{year}{2018}\natexlab{}.
\newblock \showarticletitle{Total Capture: A 3D Deformation Model for Tracking Faces, Hands, and Bodies}. In \bibinfo{booktitle}{\emph{IEEE Conf. Comput. Vis. Pattern Recog.}} \bibinfo{pages}{8320--8329}.
\newblock


\bibitem[Karaev et~al\mbox{.}(2024)]%
        {karaev2024cotracker}
\bibfield{author}{\bibinfo{person}{Nikita Karaev}, \bibinfo{person}{Ignacio Rocco}, \bibinfo{person}{Benjamin Graham}, \bibinfo{person}{Natalia Neverova}, \bibinfo{person}{Andrea Vedaldi}, {and} \bibinfo{person}{Christian Rupprecht}.} \bibinfo{year}{2024}\natexlab{}.
\newblock \showarticletitle{Cotracker: It is better to track together}. In \bibinfo{booktitle}{\emph{European Conference on Computer Vision}}. Springer, \bibinfo{pages}{18--35}.
\newblock


\bibitem[Kavan et~al\mbox{.}(2007)]%
        {kavan2007skinning}
\bibfield{author}{\bibinfo{person}{Ladislav Kavan}, \bibinfo{person}{Steven Collins}, \bibinfo{person}{Ji{\v{r}}{\'\i} {\v{Z}}{\'a}ra}, {and} \bibinfo{person}{Carol O'Sullivan}.} \bibinfo{year}{2007}\natexlab{}.
\newblock \showarticletitle{Skinning with dual quaternions}. In \bibinfo{booktitle}{\emph{Proceedings of the 2007 symposium on Interactive 3D graphics and games}}. \bibinfo{pages}{39--46}.
\newblock


\bibitem[Kerbl et~al\mbox{.}(2023)]%
        {kerbl20233d}
\bibfield{author}{\bibinfo{person}{Bernhard Kerbl}, \bibinfo{person}{Georgios Kopanas}, \bibinfo{person}{Thomas Leimk{\"u}hler}, {and} \bibinfo{person}{George Drettakis}.} \bibinfo{year}{2023}\natexlab{}.
\newblock \showarticletitle{3d gaussian splatting for real-time radiance field rendering}.
\newblock \bibinfo{journal}{\emph{ACM Trans. Graph.}} \bibinfo{volume}{42}, \bibinfo{number}{4} (\bibinfo{year}{2023}), \bibinfo{pages}{1--14}.
\newblock


\bibitem[Khirodkar et~al\mbox{.}(2024)]%
        {khirodkar2024sapiens}
\bibfield{author}{\bibinfo{person}{Rawal Khirodkar}, \bibinfo{person}{Timur Bagautdinov}, \bibinfo{person}{Julieta Martinez}, \bibinfo{person}{Su Zhaoen}, \bibinfo{person}{Austin James}, \bibinfo{person}{Peter Selednik}, \bibinfo{person}{Stuart Anderson}, {and} \bibinfo{person}{Shunsuke Saito}.} \bibinfo{year}{2024}\natexlab{}.
\newblock \showarticletitle{Sapiens: Foundation for human vision models}. In \bibinfo{booktitle}{\emph{European Conference on Computer Vision}}. Springer, \bibinfo{pages}{206--228}.
\newblock


\bibitem[Kingma and Ba(2017)]%
        {kingma2017adam}
\bibfield{author}{\bibinfo{person}{Diederik~P. Kingma} {and} \bibinfo{person}{Jimmy Ba}.} \bibinfo{year}{2017}\natexlab{}.
\newblock \bibinfo{title}{Adam: A Method for Stochastic Optimization}.
\newblock
\showeprint[arxiv]{1412.6980}~[cs.LG]


\bibitem[Kocabas et~al\mbox{.}(2024)]%
        {kocabas2023hugs}
\bibfield{author}{\bibinfo{person}{Muhammed Kocabas}, \bibinfo{person}{Jen-Hao~Rick Chang}, \bibinfo{person}{James Gabriel}, \bibinfo{person}{Oncel Tuzel}, {and} \bibinfo{person}{Anurag Ranjan}.} \bibinfo{year}{2024}\natexlab{}.
\newblock \showarticletitle{Hugs: Human gaussian splats}. In \bibinfo{booktitle}{\emph{CVPR}}.
\newblock


\bibitem[Kwon et~al\mbox{.}(2021)]%
        {kwon2021neural}
\bibfield{author}{\bibinfo{person}{Youngjoong Kwon}, \bibinfo{person}{Dahun Kim}, \bibinfo{person}{Duygu Ceylan}, {and} \bibinfo{person}{Henry Fuchs}.} \bibinfo{year}{2021}\natexlab{}.
\newblock \showarticletitle{Neural Human Performer: Learning Generalizable Radiance Fields for Human Performance Rendering}.
\newblock \bibinfo{journal}{\emph{Adv. Neural Inform. Process. Syst.}} (\bibinfo{year}{2021}).
\newblock


\bibitem[Kwon et~al\mbox{.}(2023)]%
        {kwon2023deliffas}
\bibfield{author}{\bibinfo{person}{Youngjoong Kwon}, \bibinfo{person}{Lingjie Liu}, \bibinfo{person}{Henry Fuchs}, \bibinfo{person}{Marc Habermann}, {and} \bibinfo{person}{Christian Theobalt}.} \bibinfo{year}{2023}\natexlab{}.
\newblock \showarticletitle{DELIFFAS: Deformable Light Fields for Fast Avatar Synthesis}.
\newblock \bibinfo{journal}{\emph{Adv. Neural Inform. Process. Syst.}} (\bibinfo{year}{2023}).
\newblock


\bibitem[Laine et~al\mbox{.}(2020)]%
        {Laine2020diffrast}
\bibfield{author}{\bibinfo{person}{Samuli Laine}, \bibinfo{person}{Janne Hellsten}, \bibinfo{person}{Tero Karras}, \bibinfo{person}{Yeongho Seol}, \bibinfo{person}{Jaakko Lehtinen}, {and} \bibinfo{person}{Timo Aila}.} \bibinfo{year}{2020}\natexlab{}.
\newblock \showarticletitle{Modular Primitives for High-Performance Differentiable Rendering}.
\newblock \bibinfo{journal}{\emph{ACM Transactions on Graphics}} \bibinfo{volume}{39}, \bibinfo{number}{6} (\bibinfo{year}{2020}).
\newblock


\bibitem[Lei et~al\mbox{.}(2024)]%
        {lei2023gart}
\bibfield{author}{\bibinfo{person}{Jiahui Lei}, \bibinfo{person}{Yufu Wang}, \bibinfo{person}{Georgios Pavlakos}, \bibinfo{person}{Lingjie Liu}, {and} \bibinfo{person}{Kostas Daniilidis}.} \bibinfo{year}{2024}\natexlab{}.
\newblock \showarticletitle{Gart: Gaussian articulated template models}. In \bibinfo{booktitle}{\emph{CVPR}}.
\newblock


\bibitem[Li et~al\mbox{.}(2022)]%
        {li2022tava}
\bibfield{author}{\bibinfo{person}{Ruilong Li}, \bibinfo{person}{Julian Tanke}, \bibinfo{person}{Minh Vo}, \bibinfo{person}{Michael Zollhofer}, \bibinfo{person}{Jurgen Gall}, \bibinfo{person}{Angjoo Kanazawa}, {and} \bibinfo{person}{Christoph Lassner}.} \bibinfo{year}{2022}\natexlab{}.
\newblock \showarticletitle{TAVA: Template-free animatable volumetric actors}.
\newblock \bibinfo{journal}{\emph{Eur. Conf. Comput. Vis.}}
\newblock


\bibitem[Li et~al\mbox{.}(2024)]%
        {li2024animatable}
\bibfield{author}{\bibinfo{person}{Zhe Li}, \bibinfo{person}{Zerong Zheng}, \bibinfo{person}{Lizhen Wang}, {and} \bibinfo{person}{Yebin Liu}.} \bibinfo{year}{2024}\natexlab{}.
\newblock \showarticletitle{Animatable gaussians: Learning pose-dependent gaussian maps for high-fidelity human avatar modeling}. In \bibinfo{booktitle}{\emph{Proceedings of the IEEE/CVF conference on computer vision and pattern recognition}}. \bibinfo{pages}{19711--19722}.
\newblock


\bibitem[Liang et~al\mbox{.}(2024)]%
        {liang2024analyticsplatting}
\bibfield{author}{\bibinfo{person}{Zhihao Liang}, \bibinfo{person}{Qi Zhang}, \bibinfo{person}{Wenbo Hu}, \bibinfo{person}{Ying Feng}, \bibinfo{person}{Lei Zhu}, {and} \bibinfo{person}{Kui Jia}.} \bibinfo{year}{2024}\natexlab{}.
\newblock \bibinfo{title}{Analytic-Splatting: Anti-Aliased 3D Gaussian Splatting via Analytic Integration}.
\newblock
\showeprint[arxiv]{2403.11056}~[cs.CV]


\bibitem[Lin et~al\mbox{.}(2022)]%
        {lin2022learning}
\bibfield{author}{\bibinfo{person}{Siyou Lin}, \bibinfo{person}{Hongwen Zhang}, \bibinfo{person}{Zerong Zheng}, \bibinfo{person}{Ruizhi Shao}, {and} \bibinfo{person}{Yebin Liu}.} \bibinfo{year}{2022}\natexlab{}.
\newblock \showarticletitle{Learning implicit templates for point-based clothed human modeling}. In \bibinfo{booktitle}{\emph{ECCV}}. Springer, \bibinfo{pages}{210--228}.
\newblock


\bibitem[Liu et~al\mbox{.}(2021)]%
        {liu2021neural}
\bibfield{author}{\bibinfo{person}{Lingjie Liu}, \bibinfo{person}{Marc Habermann}, \bibinfo{person}{Viktor Rudnev}, \bibinfo{person}{Kripasindhu Sarkar}, \bibinfo{person}{Jiatao Gu}, {and} \bibinfo{person}{Christian Theobalt}.} \bibinfo{year}{2021}\natexlab{}.
\newblock \showarticletitle{Neural Actor: Neural Free-view Synthesis of Human Actors with Pose Control}.
\newblock \bibinfo{journal}{\emph{ACM Trans. Graph.(ACM SIGGRAPH Asia)}} (\bibinfo{year}{2021}).
\newblock


\bibitem[Lombardi et~al\mbox{.}(2021)]%
        {Lombardi2021MVP}
\bibfield{author}{\bibinfo{person}{Stephen Lombardi}, \bibinfo{person}{Tomas Simon}, \bibinfo{person}{Gabriel Schwartz}, \bibinfo{person}{Michael Zollhofer}, \bibinfo{person}{Yaser Sheikh}, {and} \bibinfo{person}{Jason~M. Saragih}.} \bibinfo{year}{2021}\natexlab{}.
\newblock \showarticletitle{Mixture of volumetric primitives for efficient neural rendering}.
\newblock \bibinfo{journal}{\emph{ACM Trans. Graph.}} \bibinfo{volume}{40}, \bibinfo{number}{4} (\bibinfo{year}{2021}), \bibinfo{pages}{59:1--59:13}.
\newblock


\bibitem[Loper et~al\mbox{.}(2015a)]%
        {loper15}
\bibfield{author}{\bibinfo{person}{Matthew Loper}, \bibinfo{person}{Naureen Mahmood}, \bibinfo{person}{Javier Romero}, \bibinfo{person}{Gerard Pons-Moll}, {and} \bibinfo{person}{Michael~J. Black}.} \bibinfo{year}{2015}\natexlab{a}.
\newblock \showarticletitle{{SMPL}: A Skinned Multi-Person Linear Model}.
\newblock \bibinfo{journal}{\emph{ACM Trans. Graphics (Proc. SIGGRAPH Asia)}} \bibinfo{volume}{34}, \bibinfo{number}{6} (\bibinfo{date}{Oct} \bibinfo{year}{2015}), \bibinfo{pages}{248:1--248:16}.
\newblock


\bibitem[Loper et~al\mbox{.}(2015b)]%
        {loper2015smpl}
\bibfield{author}{\bibinfo{person}{Matthew Loper}, \bibinfo{person}{Naureen Mahmood}, \bibinfo{person}{Javier Romero}, \bibinfo{person}{Gerard Pons-Moll}, {and} \bibinfo{person}{Michael~J Black}.} \bibinfo{year}{2015}\natexlab{b}.
\newblock \showarticletitle{SMPL: A Skinned Multi-Person Linear Model}.
\newblock \bibinfo{journal}{\emph{ACM Transactions on Graphics}} \bibinfo{volume}{34}, \bibinfo{number}{6} (\bibinfo{year}{2015}).
\newblock


\bibitem[Ma et~al\mbox{.}(2021a)]%
        {ma2021scale}
\bibfield{author}{\bibinfo{person}{Qianli Ma}, \bibinfo{person}{Shunsuke Saito}, \bibinfo{person}{Jinlong Yang}, \bibinfo{person}{Siyu Tang}, {and} \bibinfo{person}{Michael~J Black}.} \bibinfo{year}{2021}\natexlab{a}.
\newblock \showarticletitle{SCALE: Modeling clothed humans with a surface codec of articulated local elements}. In \bibinfo{booktitle}{\emph{CVPR}}. \bibinfo{pages}{16082--16093}.
\newblock


\bibitem[Ma et~al\mbox{.}(2021b)]%
        {ma2021power}
\bibfield{author}{\bibinfo{person}{Qianli Ma}, \bibinfo{person}{Jinlong Yang}, \bibinfo{person}{Siyu Tang}, {and} \bibinfo{person}{Michael~J Black}.} \bibinfo{year}{2021}\natexlab{b}.
\newblock \showarticletitle{The power of points for modeling humans in clothing}. In \bibinfo{booktitle}{\emph{ICCV}}. \bibinfo{pages}{10974--10984}.
\newblock


\bibitem[Ma{\'c}kiewicz and Ratajczak(1993)]%
        {mackiewicz1993principal}
\bibfield{author}{\bibinfo{person}{Andrzej Ma{\'c}kiewicz} {and} \bibinfo{person}{Waldemar Ratajczak}.} \bibinfo{year}{1993}\natexlab{}.
\newblock \showarticletitle{Principal components analysis (PCA)}.
\newblock \bibinfo{journal}{\emph{Computers \& Geosciences}} \bibinfo{volume}{19}, \bibinfo{number}{3} (\bibinfo{year}{1993}), \bibinfo{pages}{303--342}.
\newblock


\bibitem[Mildenhall et~al\mbox{.}(2020)]%
        {mildenhall2020nerf}
\bibfield{author}{\bibinfo{person}{Ben Mildenhall}, \bibinfo{person}{Pratul~P. Srinivasan}, \bibinfo{person}{Matthew Tancik}, \bibinfo{person}{Jonathan~T. Barron}, \bibinfo{person}{Ravi Ramamoorthi}, {and} \bibinfo{person}{Ren Ng}.} \bibinfo{year}{2020}\natexlab{}.
\newblock \showarticletitle{NeRF: Representing Scenes as Neural Radiance Fields for View Synthesis}. In \bibinfo{booktitle}{\emph{Eur. Conf. Comput. Vis.}}
\newblock


\bibitem[Osman et~al\mbox{.}(2020)]%
        {STAR:2020}
\bibfield{author}{\bibinfo{person}{A. Osman}, \bibinfo{person}{Timo Bolkart}, {and} \bibinfo{person}{Michael~J. Black}.} \bibinfo{year}{2020}\natexlab{}.
\newblock \showarticletitle{STAR: Sparse Trained Articulated Human Body Regressor}. In \bibinfo{booktitle}{\emph{Eur. Conf. Comput. Vis.}} \bibinfo{pages}{598--613}.
\newblock


\bibitem[Pang et~al\mbox{.}(2024)]%
        {Pang_2024_CVPR}
\bibfield{author}{\bibinfo{person}{Haokai Pang}, \bibinfo{person}{Heming Zhu}, \bibinfo{person}{Adam Kortylewski}, \bibinfo{person}{Christian Theobalt}, {and} \bibinfo{person}{Marc Habermann}.} \bibinfo{year}{2024}\natexlab{}.
\newblock \showarticletitle{ASH: Animatable Gaussian Splats for Efficient and Photoreal Human Rendering}. In \bibinfo{booktitle}{\emph{IEEE Conf. Comput. Vis. Pattern Recog.}} \bibinfo{pages}{1165--1175}.
\newblock


\bibitem[Paszke et~al\mbox{.}(2017)]%
        {paszke2017automatic}
\bibfield{author}{\bibinfo{person}{Adam Paszke}, \bibinfo{person}{Sam Gross}, \bibinfo{person}{Soumith Chintala}, \bibinfo{person}{Gregory Chanan}, \bibinfo{person}{Edward Yang}, \bibinfo{person}{Zachary DeVito}, \bibinfo{person}{Zeming Lin}, \bibinfo{person}{Alban Desmaison}, \bibinfo{person}{Luca Antiga}, {and} \bibinfo{person}{Adam Lerer}.} \bibinfo{year}{2017}\natexlab{}.
\newblock \showarticletitle{Automatic differentiation in PyTorch}. In \bibinfo{booktitle}{\emph{NIPS-W}}.
\newblock


\bibitem[Pavlakos et~al\mbox{.}(2019)]%
        {SMPL-X:2019}
\bibfield{author}{\bibinfo{person}{Georgios Pavlakos}, \bibinfo{person}{Vasileios Choutas}, \bibinfo{person}{Nima Ghorbani}, \bibinfo{person}{Timo Bolkart}, \bibinfo{person}{Ahmed A.~A. Osman}, \bibinfo{person}{Dimitrios Tzionas}, {and} \bibinfo{person}{Michael~J. Black}.} \bibinfo{year}{2019}\natexlab{}.
\newblock \showarticletitle{Expressive Body Capture: 3D Hands, Face, and Body From a Single Image}. In \bibinfo{booktitle}{\emph{IEEE Conf. Comput. Vis. Pattern Recog.}} \bibinfo{pages}{10975--10985}.
\newblock


\bibitem[Peng et~al\mbox{.}(2021a)]%
        {peng2021animatable}
\bibfield{author}{\bibinfo{person}{Sida Peng}, \bibinfo{person}{Junting Dong}, \bibinfo{person}{Qianqian Wang}, \bibinfo{person}{Shangzhan Zhang}, \bibinfo{person}{Qing Shuai}, \bibinfo{person}{Xiaowei Zhou}, {and} \bibinfo{person}{Hujun Bao}.} \bibinfo{year}{2021}\natexlab{a}.
\newblock \showarticletitle{Animatable neural radiance fields for modeling dynamic human bodies}. In \bibinfo{booktitle}{\emph{Int. Conf. Comput. Vis.}} \bibinfo{pages}{14314--14323}.
\newblock


\bibitem[Peng et~al\mbox{.}(2021b)]%
        {peng2021neuralbody}
\bibfield{author}{\bibinfo{person}{Sida Peng}, \bibinfo{person}{Yuanqing Zhang}, \bibinfo{person}{Yinghao Xu}, \bibinfo{person}{Qianqian Wang}, \bibinfo{person}{Qing Shuai}, \bibinfo{person}{Hujun Bao}, {and} \bibinfo{person}{Xiaowei Zhou}.} \bibinfo{year}{2021}\natexlab{b}.
\newblock \showarticletitle{Neural Body: Implicit Neural Representations With Structured Latent Codes for Novel View Synthesis of Dynamic Humans}. In \bibinfo{booktitle}{\emph{IEEE Conf. Comput. Vis. Pattern Recog.}} \bibinfo{pages}{9054--9063}.
\newblock


\bibitem[Qian et~al\mbox{.}(2024)]%
        {qian20233dgs}
\bibfield{author}{\bibinfo{person}{Zhiyin Qian}, \bibinfo{person}{Shaofei Wang}, \bibinfo{person}{Marko Mihajlovic}, \bibinfo{person}{Andreas Geiger}, {and} \bibinfo{person}{Siyu Tang}.} \bibinfo{year}{2024}\natexlab{}.
\newblock \showarticletitle{3DGS-Avatar: Animatable Avatars via Deformable 3D Gaussian Splatting}. In \bibinfo{booktitle}{\emph{CVPR}}.
\newblock


\bibitem[Remelli et~al\mbox{.}(2022)]%
        {Remelli2022TexelAligned}
\bibfield{author}{\bibinfo{person}{Edoardo Remelli}, \bibinfo{person}{Timur~M. Bagautdinov}, \bibinfo{person}{Shunsuke Saito}, \bibinfo{person}{Chenglei Wu}, \bibinfo{person}{Tomas Simon}, \bibinfo{person}{Shih{-}En Wei}, \bibinfo{person}{Kaiwen Guo}, \bibinfo{person}{Zhe Cao}, \bibinfo{person}{Fabian Prada}, \bibinfo{person}{Jason~M. Saragih}, {and} \bibinfo{person}{Yaser Sheikh}.} \bibinfo{year}{2022}\natexlab{}.
\newblock \showarticletitle{Drivable Volumetric Avatars using Texel-Aligned Features}. In \bibinfo{booktitle}{\emph{SIGGRAPH (Conference Paper Track)}}. \bibinfo{pages}{56:1--56:9}.
\newblock


\bibitem[Shetty et~al\mbox{.}(2024)]%
        {shetty2023holoported}
\bibfield{author}{\bibinfo{person}{Ashwath Shetty}, \bibinfo{person}{Marc Habermann}, \bibinfo{person}{Guoxing Sun}, \bibinfo{person}{Diogo Luvizon}, \bibinfo{person}{Vladislav Golyanik}, {and} \bibinfo{person}{Christian Theobalt}.} \bibinfo{year}{2024}\natexlab{}.
\newblock \showarticletitle{Holoported characters: Real-time free-viewpoint rendering of humans from sparse rgb cameras}. In \bibinfo{booktitle}{\emph{Proceedings of the IEEE/CVF Conference on Computer Vision and Pattern Recognition}}. \bibinfo{pages}{1206--1215}.
\newblock


\bibitem[Shysheya et~al\mbox{.}(2019)]%
        {shysheya2019textured}
\bibfield{author}{\bibinfo{person}{Aliaksandra Shysheya}, \bibinfo{person}{Egor Zakharov}, \bibinfo{person}{Kara-Ali Aliev}, \bibinfo{person}{Renat Bashirov}, \bibinfo{person}{Egor Burkov}, \bibinfo{person}{Karim Iskakov}, \bibinfo{person}{Aleksei Ivakhnenko}, \bibinfo{person}{Yury Malkov}, \bibinfo{person}{Igor Pasechnik}, \bibinfo{person}{Dmitry Ulyanov}, {et~al\mbox{.}}} \bibinfo{year}{2019}\natexlab{}.
\newblock \showarticletitle{Textured neural avatars}. In \bibinfo{booktitle}{\emph{IEEE Conf. Comput. Vis. Pattern Recog.}} \bibinfo{pages}{2387--2397}.
\newblock


\bibitem[Stoll et~al\mbox{.}(2010)]%
        {stoll2010video}
\bibfield{author}{\bibinfo{person}{Carsten Stoll}, \bibinfo{person}{Juergen Gall}, \bibinfo{person}{Edilson De~Aguiar}, \bibinfo{person}{Sebastian Thrun}, {and} \bibinfo{person}{Christian Theobalt}.} \bibinfo{year}{2010}\natexlab{}.
\newblock \showarticletitle{Video-based reconstruction of animatable human characters}.
\newblock \bibinfo{journal}{\emph{TOG}} \bibinfo{volume}{29}, \bibinfo{number}{6} (\bibinfo{year}{2010}), \bibinfo{pages}{1--10}.
\newblock


\bibitem[Su et~al\mbox{.}(2021)]%
        {su2021nerf}
\bibfield{author}{\bibinfo{person}{Shih-Yang Su}, \bibinfo{person}{Frank Yu}, \bibinfo{person}{Michael Zollhofer}, {and} \bibinfo{person}{Helge Rhodin}.} \bibinfo{year}{2021}\natexlab{}.
\newblock \showarticletitle{A-nerf: Articulated neural radiance fields for learning human shape, appearance, and pose}.
\newblock \bibinfo{journal}{\emph{Adv. Neural Inform. Process. Syst.}}  \bibinfo{volume}{34} (\bibinfo{year}{2021}), \bibinfo{pages}{12278--12291}.
\newblock


\bibitem[Sumner et~al\mbox{.}(2007)]%
        {embedded}
\bibfield{author}{\bibinfo{person}{Robert~W. Sumner}, \bibinfo{person}{Johannes Schmid}, {and} \bibinfo{person}{Mark Pauly}.} \bibinfo{year}{2007}\natexlab{}.
\newblock \showarticletitle{Embedded Deformation for Shape Manipulation}.
\newblock \bibinfo{journal}{\emph{ACM Trans. Graph.}} \bibinfo{volume}{26}, \bibinfo{number}{3} (\bibinfo{date}{jul} \bibinfo{year}{2007}), \bibinfo{pages}{80–es}.
\newblock
\showISSN{0730-0301}
\href{https://doi.org/10.1145/1276377.1276478}{doi:\nolinkurl{10.1145/1276377.1276478}}


\bibitem[Sun et~al\mbox{.}(2025)]%
        {sun2025real}
\bibfield{author}{\bibinfo{person}{Guoxing Sun}, \bibinfo{person}{Rishabh Dabral}, \bibinfo{person}{Heming Zhu}, \bibinfo{person}{Pascal Fua}, \bibinfo{person}{Christian Theobalt}, {and} \bibinfo{person}{Marc Habermann}.} \bibinfo{year}{2025}\natexlab{}.
\newblock \showarticletitle{Real-time Free-view Human Rendering from Sparse-view RGB Videos using Double Unprojected Textures}.
\newblock  (\bibinfo{date}{June} \bibinfo{year}{2025}).
\newblock


\bibitem[{TheCaptury}(2020)]%
        {captury}
\bibfield{author}{\bibinfo{person}{{TheCaptury}}.} \bibinfo{year}{2020}\natexlab{}.
\newblock \bibinfo{title}{{The Captury}}.
\newblock \bibinfo{howpublished}{\url{http://www.thecaptury.com/}}.
\newblock


\bibitem[Wang et~al\mbox{.}(2021a)]%
        {wang2021neus}
\bibfield{author}{\bibinfo{person}{Peng Wang}, \bibinfo{person}{Lingjie Liu}, \bibinfo{person}{Yuan Liu}, \bibinfo{person}{Christian Theobalt}, \bibinfo{person}{Taku Komura}, {and} \bibinfo{person}{Wenping Wang}.} \bibinfo{year}{2021}\natexlab{a}.
\newblock \showarticletitle{NeuS: learning neural implicit surfaces by volume rendering for multi-view reconstruction}. In \bibinfo{booktitle}{\emph{Proceedings of the 35th International Conference on Neural Information Processing Systems}}. \bibinfo{pages}{27171--27183}.
\newblock


\bibitem[Wang et~al\mbox{.}(2021b)]%
        {wang2021ibrnet}
\bibfield{author}{\bibinfo{person}{Qianqian Wang}, \bibinfo{person}{Zhicheng Wang}, \bibinfo{person}{Kyle Genova}, \bibinfo{person}{Pratul Srinivasan}, \bibinfo{person}{Howard Zhou}, \bibinfo{person}{Jonathan~T. Barron}, \bibinfo{person}{Ricardo Martin-Brualla}, \bibinfo{person}{Noah Snavely}, {and} \bibinfo{person}{Thomas Funkhouser}.} \bibinfo{year}{2021}\natexlab{b}.
\newblock \showarticletitle{IBRNet: Learning Multi-View Image-Based Rendering}. In \bibinfo{booktitle}{\emph{IEEE Conf. Comput. Vis. Pattern Recog.}}
\newblock


\bibitem[Wang et~al\mbox{.}(2024)]%
        {wang2024survey}
\bibfield{author}{\bibinfo{person}{Ruihe Wang}, \bibinfo{person}{Yukang Cao}, \bibinfo{person}{Kai Han}, {and} \bibinfo{person}{Kwan-Yee~K Wong}.} \bibinfo{year}{2024}\natexlab{}.
\newblock \showarticletitle{A Survey on 3D Human Avatar Modeling--From Reconstruction to Generation}.
\newblock \bibinfo{journal}{\emph{arXiv preprint arXiv:2406.04253}} (\bibinfo{year}{2024}).
\newblock


\bibitem[Wang et~al\mbox{.}(2022)]%
        {ARAH}
\bibfield{author}{\bibinfo{person}{Shaofei Wang}, \bibinfo{person}{Katja Schwarz}, \bibinfo{person}{Andreas Geiger}, {and} \bibinfo{person}{Siyu Tang}.} \bibinfo{year}{2022}\natexlab{}.
\newblock \showarticletitle{ARAH: Animatable Volume Rendering of Articulated Human SDFs}. In \bibinfo{booktitle}{\emph{Eur. Conf. Comput. Vis.}}
\newblock


\bibitem[Wang et~al\mbox{.}(2023)]%
        {neus2}
\bibfield{author}{\bibinfo{person}{Yiming Wang}, \bibinfo{person}{Qin Han}, \bibinfo{person}{Marc Habermann}, \bibinfo{person}{Kostas Daniilidis}, \bibinfo{person}{Christian Theobalt}, {and} \bibinfo{person}{Lingjie Liu}.} \bibinfo{year}{2023}\natexlab{}.
\newblock \showarticletitle{NeuS2: Fast Learning of Neural Implicit Surfaces for Multi-view Reconstruction}. In \bibinfo{booktitle}{\emph{Int. Conf. Comput. Vis.}}
\newblock


\bibitem[Wang et~al\mbox{.}(2018)]%
        {wang2018image}
\bibfield{author}{\bibinfo{person}{Yi Wang}, \bibinfo{person}{Xin Tao}, \bibinfo{person}{Xiaojuan Qi}, \bibinfo{person}{Xiaoyong Shen}, {and} \bibinfo{person}{Jiaya Jia}.} \bibinfo{year}{2018}\natexlab{}.
\newblock \showarticletitle{Image inpainting via generative multi-column convolutional neural networks}.
\newblock \bibinfo{journal}{\emph{Advances in neural information processing systems}}  \bibinfo{volume}{31} (\bibinfo{year}{2018}).
\newblock


\bibitem[Wang et~al\mbox{.}(2020)]%
        {wang2020learning}
\bibfield{author}{\bibinfo{person}{Ziyan Wang}, \bibinfo{person}{Timur Bagautdinov}, \bibinfo{person}{Stephen Lombardi}, \bibinfo{person}{Tomas Simon}, \bibinfo{person}{Jason Saragih}, \bibinfo{person}{Jessica Hodgins}, {and} \bibinfo{person}{Michael Zollhofer}.} \bibinfo{year}{2020}\natexlab{}.
\newblock \bibinfo{title}{Learning Compositional Radiance Fields of Dynamic Human Heads}.
\newblock
\showeprint[arxiv]{2012.09955}~[cs.CV]


\bibitem[Weng et~al\mbox{.}(2022)]%
        {weng_humannerf_2022_cvpr}
\bibfield{author}{\bibinfo{person}{Chung-Yi Weng}, \bibinfo{person}{Brian Curless}, \bibinfo{person}{Pratul~P. Srinivasan}, \bibinfo{person}{Jonathan~T. Barron}, {and} \bibinfo{person}{Ira Kemelmacher-Shlizerman}.} \bibinfo{year}{2022}\natexlab{}.
\newblock \showarticletitle{Human{N}e{RF}: Free-Viewpoint Rendering of Moving People From Monocular Video}. In \bibinfo{booktitle}{\emph{IEEE Conf. Comput. Vis. Pattern Recog.}} \bibinfo{pages}{16210--16220}.
\newblock


\bibitem[Xiang et~al\mbox{.}(2022)]%
        {xiang2022dressing}
\bibfield{author}{\bibinfo{person}{Donglai Xiang}, \bibinfo{person}{Timur Bagautdinov}, \bibinfo{person}{Tuur Stuyck}, \bibinfo{person}{Fabian Prada}, \bibinfo{person}{Javier Romero}, \bibinfo{person}{Weipeng Xu}, \bibinfo{person}{Shunsuke Saito}, \bibinfo{person}{Jingfan Guo}, \bibinfo{person}{Breannan Smith}, \bibinfo{person}{Takaaki Shiratori}, {et~al\mbox{.}}} \bibinfo{year}{2022}\natexlab{}.
\newblock \showarticletitle{Dressing avatars: Deep photorealistic appearance for physically simulated clothing}.
\newblock \bibinfo{journal}{\emph{ACM Trans. Graph.}} \bibinfo{volume}{41}, \bibinfo{number}{6} (\bibinfo{year}{2022}), \bibinfo{pages}{1--15}.
\newblock


\bibitem[Xiang et~al\mbox{.}(2021)]%
        {xiang2021modeling}
\bibfield{author}{\bibinfo{person}{Donglai Xiang}, \bibinfo{person}{Fabian Prada}, \bibinfo{person}{Timur Bagautdinov}, \bibinfo{person}{Weipeng Xu}, \bibinfo{person}{Yuan Dong}, \bibinfo{person}{He Wen}, \bibinfo{person}{Jessica Hodgins}, {and} \bibinfo{person}{Chenglei Wu}.} \bibinfo{year}{2021}\natexlab{}.
\newblock \showarticletitle{Modeling clothing as a separate layer for an animatable human avatar}.
\newblock \bibinfo{journal}{\emph{ACM Trans. Graph.}} \bibinfo{volume}{40}, \bibinfo{number}{6} (\bibinfo{year}{2021}), \bibinfo{pages}{1--15}.
\newblock


\bibitem[Xiu et~al\mbox{.}(2023)]%
        {xiu2023econ}
\bibfield{author}{\bibinfo{person}{Yuliang Xiu}, \bibinfo{person}{Jinlong Yang}, \bibinfo{person}{Xu Cao}, \bibinfo{person}{Dimitrios Tzionas}, {and} \bibinfo{person}{Michael~J. Black}.} \bibinfo{year}{2023}\natexlab{}.
\newblock \showarticletitle{{ECON: Explicit Clothed humans Optimized via Normal integration}}. In \bibinfo{booktitle}{\emph{IEEE Conf. Comput. Vis. Pattern Recog.}}
\newblock


\bibitem[Xu et~al\mbox{.}(2011)]%
        {xu2011video}
\bibfield{author}{\bibinfo{person}{Feng Xu}, \bibinfo{person}{Yebin Liu}, \bibinfo{person}{Carsten Stoll}, \bibinfo{person}{James Tompkin}, \bibinfo{person}{Gaurav Bharaj}, \bibinfo{person}{Qionghai Dai}, \bibinfo{person}{Hans-Peter Seidel}, \bibinfo{person}{Jan Kautz}, {and} \bibinfo{person}{Christian Theobalt}.} \bibinfo{year}{2011}\natexlab{}.
\newblock \showarticletitle{Video-based characters: creating new human performances from a multi-view video database}.
\newblock In \bibinfo{booktitle}{\emph{ACM SIGGRAPH 2011 papers}}. \bibinfo{pages}{1--10}.
\newblock


\bibitem[Xu et~al\mbox{.}(2021)]%
        {xu2021hnerf}
\bibfield{author}{\bibinfo{person}{Hongyi Xu}, \bibinfo{person}{Thiemo Alldieck}, {and} \bibinfo{person}{Cristian Sminchisescu}.} \bibinfo{year}{2021}\natexlab{}.
\newblock \showarticletitle{H-nerf: Neural radiance fields for rendering and temporal reconstruction of humans in motion}.
\newblock \bibinfo{journal}{\emph{Adv. Neural Inform. Process. Syst.}}  \bibinfo{volume}{34} (\bibinfo{year}{2021}), \bibinfo{pages}{14955--14966}.
\newblock


\bibitem[Xu et~al\mbox{.}(2024)]%
        {xu2024representing}
\bibfield{author}{\bibinfo{person}{Zhen Xu}, \bibinfo{person}{Yinghao Xu}, \bibinfo{person}{Zhiyuan Yu}, \bibinfo{person}{Sida Peng}, \bibinfo{person}{Jiaming Sun}, \bibinfo{person}{Hujun Bao}, {and} \bibinfo{person}{Xiaowei Zhou}.} \bibinfo{year}{2024}\natexlab{}.
\newblock \showarticletitle{Representing long volumetric video with temporal gaussian hierarchy}.
\newblock \bibinfo{journal}{\emph{ACM Transactions on Graphics (TOG)}} \bibinfo{volume}{43}, \bibinfo{number}{6} (\bibinfo{year}{2024}), \bibinfo{pages}{1--18}.
\newblock


\bibitem[Ye et~al\mbox{.}(2025)]%
        {ye2025gsplat}
\bibfield{author}{\bibinfo{person}{Vickie Ye}, \bibinfo{person}{Ruilong Li}, \bibinfo{person}{Justin Kerr}, \bibinfo{person}{Matias Turkulainen}, \bibinfo{person}{Brent Yi}, \bibinfo{person}{Zhuoyang Pan}, \bibinfo{person}{Otto Seiskari}, \bibinfo{person}{Jianbo Ye}, \bibinfo{person}{Jeffrey Hu}, \bibinfo{person}{Matthew Tancik}, {and} \bibinfo{person}{Angjoo Kanazawa}.} \bibinfo{year}{2025}\natexlab{}.
\newblock \showarticletitle{gsplat: An open-source library for Gaussian splatting}.
\newblock \bibinfo{journal}{\emph{Journal of Machine Learning Research}} \bibinfo{volume}{26}, \bibinfo{number}{34} (\bibinfo{year}{2025}), \bibinfo{pages}{1--17}.
\newblock


\bibitem[Yi et~al\mbox{.}(2025)]%
        {yi2025viser}
\bibfield{author}{\bibinfo{person}{Brent Yi}, \bibinfo{person}{Chung~Min Kim}, \bibinfo{person}{Justin Kerr}, \bibinfo{person}{Gina Wu}, \bibinfo{person}{Rebecca Feng}, \bibinfo{person}{Anthony Zhang}, \bibinfo{person}{Jonas Kulhanek}, \bibinfo{person}{Hongsuk Choi}, \bibinfo{person}{Yi Ma}, \bibinfo{person}{Matthew Tancik}, {and} \bibinfo{person}{Angjoo Kanazawa}.} \bibinfo{year}{2025}\natexlab{}.
\newblock \bibinfo{title}{Viser: Imperative, Web-based 3D Visualization in Python}.
\newblock
\showeprint[arxiv]{2507.22885}~[cs.CV]
\urldef\tempurl%
\url{https://arxiv.org/abs/2507.22885}
\showURL{%
\tempurl}


\bibitem[Zhang et~al\mbox{.}(2018)]%
        {zhang2018perceptual}
\bibfield{author}{\bibinfo{person}{Richard Zhang}, \bibinfo{person}{Phillip Isola}, \bibinfo{person}{Alexei~A Efros}, \bibinfo{person}{Eli Shechtman}, {and} \bibinfo{person}{Oliver Wang}.} \bibinfo{year}{2018}\natexlab{}.
\newblock \showarticletitle{The Unreasonable Effectiveness of Deep Features as a Perceptual Metric}. In \bibinfo{booktitle}{\emph{IEEE Conf. Comput. Vis. Pattern Recog.}}
\newblock


\bibitem[Zhang et~al\mbox{.}(2024)]%
        {zhang2024sifu}
\bibfield{author}{\bibinfo{person}{Zechuan Zhang}, \bibinfo{person}{Zongxin Yang}, {and} \bibinfo{person}{Yi Yang}.} \bibinfo{year}{2024}\natexlab{}.
\newblock \showarticletitle{Sifu: Side-view conditioned implicit function for real-world usable clothed human reconstruction}. In \bibinfo{booktitle}{\emph{IEEE Conf. Comput. Vis. Pattern Recog.}} \bibinfo{pages}{9936--9947}.
\newblock


\bibitem[Zheng et~al\mbox{.}(2025)]%
        {zheng2025gstar}
\bibfield{author}{\bibinfo{person}{Chengwei Zheng}, \bibinfo{person}{Lixin Xue}, \bibinfo{person}{Juan Zarate}, {and} \bibinfo{person}{Jie Song}.} \bibinfo{year}{2025}\natexlab{}.
\newblock \showarticletitle{GSTAR: Gaussian Surface Tracking and Reconstruction}.
\newblock \bibinfo{journal}{\emph{arXiv preprint arXiv:2501.10283}} (\bibinfo{year}{2025}).
\newblock


\bibitem[Zheng et~al\mbox{.}(2024)]%
        {PhysAavatar24}
\bibfield{author}{\bibinfo{person}{Yang Zheng}, \bibinfo{person}{Qingqing Zhao}, \bibinfo{person}{Guandao Yang}, \bibinfo{person}{Wang Yifan}, \bibinfo{person}{Donglai Xiang}, \bibinfo{person}{Florian Dubost}, \bibinfo{person}{Dmitry Lagun}, \bibinfo{person}{Thabo Beeler}, \bibinfo{person}{Federico Tombari}, \bibinfo{person}{Leonidas Guibas}, {and} \bibinfo{person}{Gordon Wetzstein}.} \bibinfo{year}{2024}\natexlab{}.
\newblock \showarticletitle{PhysAvatar: Learning the Physics of Dressed 3D Avatars from Visual Observations}.
\newblock \bibinfo{journal}{\emph{European Conference on Computer Vision (ECCV)}}.
\newblock


\bibitem[Zheng et~al\mbox{.}(2023)]%
        {zheng2023avatarrex}
\bibfield{author}{\bibinfo{person}{Zerong Zheng}, \bibinfo{person}{Xiaochen Zhao}, \bibinfo{person}{Hongwen Zhang}, \bibinfo{person}{Boning Liu}, {and} \bibinfo{person}{Yebin Liu}.} \bibinfo{year}{2023}\natexlab{}.
\newblock \showarticletitle{AvatarRex: Real-time Expressive Full-body Avatars}.
\newblock \bibinfo{journal}{\emph{ACM Trans. Graph.}} \bibinfo{volume}{42}, \bibinfo{number}{4} (\bibinfo{year}{2023}).
\newblock


\bibitem[Zhu et~al\mbox{.}(2022)]%
        {zhu2022registering}
\bibfield{author}{\bibinfo{person}{Heming Zhu}, \bibinfo{person}{Lingteng Qiu}, \bibinfo{person}{Yuda Qiu}, {and} \bibinfo{person}{Xiaoguang Han}.} \bibinfo{year}{2022}\natexlab{}.
\newblock \showarticletitle{Registering explicit to implicit: Towards high-fidelity garment mesh reconstruction from single images}. In \bibinfo{booktitle}{\emph{IEEE Conf. Comput. Vis. Pattern Recog.}} \bibinfo{pages}{3845--3854}.
\newblock


\bibitem[Zhu et~al\mbox{.}(2023)]%
        {zhu2023trihuman}
\bibfield{author}{\bibinfo{person}{Heming Zhu}, \bibinfo{person}{Fangneng Zhan}, \bibinfo{person}{Christian Theobalt}, {and} \bibinfo{person}{Marc Habermann}.} \bibinfo{year}{2023}\natexlab{}.
\newblock \showarticletitle{TriHuman: A Real-time and Controllable Tri-plane Representation for Detailed Human Geometry and Appearance Synthesis}.
\newblock \bibinfo{journal}{\emph{arXiv preprint arXiv:2312.05161}} (\bibinfo{year}{2023}).
\newblock


\bibitem[Zhu et~al\mbox{.}(2017)]%
        {CycleGAN2017}
\bibfield{author}{\bibinfo{person}{Jun-Yan Zhu}, \bibinfo{person}{Taesung Park}, \bibinfo{person}{Phillip Isola}, {and} \bibinfo{person}{Alexei~A Efros}.} \bibinfo{year}{2017}\natexlab{}.
\newblock \showarticletitle{Unpaired Image-to-Image Translation using Cycle-Consistent Adversarial Networkss}. In \bibinfo{booktitle}{\emph{Computer Vision (ICCV), 2017 IEEE International Conference on}}.
\newblock


\end{thebibliography}

\appendix

%
%
\section{Gaussian Avatar Representation} \label{suppsec:1}
%
%
\par \noindent \textbf{Gaussian-texel to Mesh Conversion.}
The detailed surface geometry produced by \titleabr is represented as a mesh with consistent triangulation, where each vertex corresponds to the posed position of a Gaussian texels $\boldsymbol{\bar{\mu}}_i$.
To construct the triangulation in texel space, for each texel $(i, j)$, we form a triangle with the neighboring texels $(i + 1, j)$, $(i, j + 1)$, and $(i + 1, j + 1)$, provided they are covered by the texture.
However, due to texture seams on the template mesh, the resulting texel triangulation may also contain discontinuities along these seams.
To close the resulting gaps, we propose a simple yet effective strategy: 
1) We assume the template mesh consists of a single texture island and that its seams form a Directed Acyclic Graph (DAG).
2) For each seam endpoint, we iteratively extend the triangulation by connecting vertices across the seam with the shortest possible edge, gradually stitching the seam while minimizing new edge lengths.

%
%
\par \noindent \textbf{Spatial Mesh Regularization.}
In, both, the depth alignment and triangle alignment stage, we adopted mesh regularization terms, consisting of Laplacian loss $\mathcal{L}_\mathrm{lap}$, Laplacian smoothness term $\mathcal{L}_\mathrm{lapz}$, and face normal consistency loss $\mathcal{L}_\mathrm{norm}$ to maintain the smoothness of the template mesh $\mathbf{V}_f$  while not loosing wrinkle details:
\begin{align} 
    \mathcal{L}_\mathrm{spatial}(\mathbf{V}_f) &= \mathcal{L}_\mathrm{lap} + \mathcal{L}_\mathrm{lapz} +\mathcal{L}_\mathrm{norm} \\
    \mathcal{L}_\mathrm{lap} &= \mathbf{L}(\mathbf{V}_f) - \mathbf{L}(\tilde{\mathbf{V}_f})\\ 
    \mathcal{L}_\mathrm{lapz} &= \frac{1}{\mathbf{N}} \sum_{i=1}^{N}  \|(\mathbf{L} \mathcal{V}_f)_i\|_{2} \\
    \mathcal{L}_\mathrm{norm} &= \frac{1}{\mathrm{N}_{\mathrm{tri}}}(\sum_{i=1}^{\mathrm{N}_{\mathrm{tri}}}\frac{\sum_{j=1}^{\mathrm{N}_{\mathrm{tri},i}}(1 - \mathbf{n}_{\mathrm{tri},i} \cdot \mathbf{n}_{\mathrm{tri},i,j})}{\mathrm{N}_{\mathrm{tri},i}})
\end{align}
where $\mathbf{L}$ denotes the vertex Laplacian operator, $\tilde{\mathbf{V}_f}$ indicates the posed and skinned template mesh; $\mathbf{n}_{\mathrm{tri},i}$ and $\mathbf{n}_{\mathrm{tri},i,j}$ stands for the face normals of face $i$ the adjacent faces normals of face $i$., $\mathrm{N}_{\mathrm{tri}}$ refers to the number of the neighboring faces for face $i$.

%
\par \noindent \textbf{Spatial Gaussian Regularization.}
Since 3D Gaussian Splatting~\cite{kerbl20233d} is a highly flexible representation, the absence of proper regularization can lead to severe artifacts—such as floating or excessively large Gaussian splats—which significantly impairs the ability of Gaussian textures to capture fine details.
Therefore, we propose to constrain the canonical offset $\mathbf{\bar{d}}$ of the 3D Gaussian Splats w.r.t. the template mesh using sigmoid function $\sigma(x)$.
\begin{equation} \label{eq:spatialgaussreg}
    \mathbf{\bar{d}} = (\sigma(\bar{d}_{\mathrm{raw}}) - 0.5) * 2 l_{lim}
\end{equation}
where $\bar{d}_{\mathrm{raw}}$ denotes the predicted offset, $l_{lim}$ denote the offset limit, which is set to 0.03 empirically.
%
%
\par \noindent \textbf{Network Architecture.} For the drivable human template mesh, i.e., the embedded and per-vertex deformation network $\mathcal{F_\mathrm{eg}}$, $\mathcal{F_\mathrm{delta}}$ we adopt the implementation from \cite{habermann2021}.
However, we augment the input to both networks by channel-wise concatenating a per-frame latent code $\mathbf{Z}_f$. 
This addition accounts for the stochastic clothing dynamics that cannot be solely modeled by skeletal motion.
The appearance $\mathcal{E_\mathrm{app}}$ and $\mathcal{E_\mathrm{geo}}$ geometry decoder are two UNets following the implementation by ~\citet{Pang_2024_CVPR}.
%
%
%
\begin{figure*}[h]
    \centering
    \includegraphics[width=\linewidth]{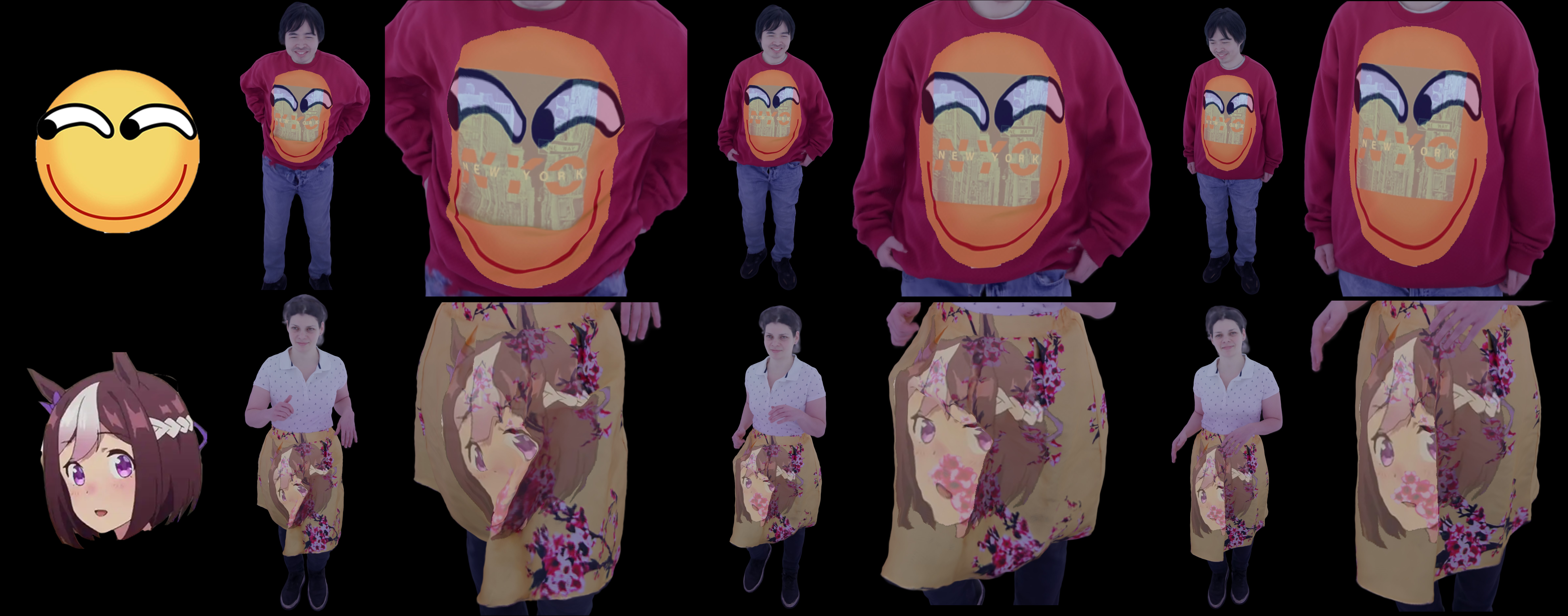}
    \vspace{-2em}
    \caption{
    \textbf{Texture editing.}
    \titleabr enables consistent texture editing.
    Given the texture image shown on the left, \titleabr seamlessly integrates the edits onto the rendered character.
    Notably, the inserted texture deforms seamlessly with the clothing wrinkles and remains consistently anchored to the character’s original texture. Please \textbf{zoom-in} to better observe the details.
    }
    \label{fig:suppltexture}
\end{figure*}
%
%

%
%
\section{Depth Alignment} \label{suppsec:3}
%
%
\par \noindent \textbf{Latent Encoder Architecture.}
For the Latent Encoder architecture, we adopt a 3 layer-MLP and the width of the MLP set to $16$.

\par \noindent \textbf{Training Details.} 
To train the depth alignment stage, the Chamfer distance loss $\mathcal{L}_{\mathrm{cham}}$, spatial regularizer $\mathcal{L}_\mathrm{lapz}$, Laplacian loss $\mathcal{L}_\mathrm{lap}$, and normal consistency loss $\mathcal{L}_\mathrm{norm}$ are assigned weights of $0.01$, $0.1$, $0.01$, and $2.5$, respectively. The depth alignment stage is trained for $360{,}000$ iterations using the Adam optimizer~\cite{kingma2017adam} with a learning rate of $1e^{-4}$ scheduled by a cosine decay, which takes around 12 hours.

Notably, since per-frame latents are unavailable at test time, we augment each training batch by sampling additional data where the network receives the skeletal pose and a zero latent as input. 
This is done alongside regular samples with learned latents to help the network adapt to the absence of per-frame latent codes during inference.
%

%
%
\section{Vertex-level Alignment} \label{suppsec:4}

\par \noindent \textbf{Training Details.} To train the depth alignent stage, the Chamfer distance loss $\mathcal{L}_{\mathrm{cham}}$, vertex alignment loss $\mathcal{L}_\mathrm{corr\text{-}vertex}$, spatial regularizer $\mathcal{L}_\mathrm{lapz}$, Laplacian loss $\mathcal{L}_\mathrm{lap}$, and normal consistency loss $\mathcal{L}_\mathrm{norm}$ are assigned weights of $0.01$, $0.02$, $0.1$, $0.01$, and $2.5$, respectively. 
The vertex-level alignment stage is trained for $360{,}000$ iterations using Adam optimizer with a learning rate of $5e^{-4}$ scheduled with a cosine decay scheduler, which takes around 12 hours.

Simar to the depth level alignment, we augment each training batch by sampling additional data where the network receives the skeletal pose and a zero latent as input. 

%
%
\section{Texel-level Alignment} \label{suppsec:5}

\par \noindent \textbf{Training Details.} To train the animatable
Gaussian texture with texel-level alignment, following the open-
sourced implementation in  \citet{Pang_2024_CVPR}, which includes 15,000
iterations of initialization before the main training, while the main
training lasts for 2,000,000 iterations with a learning rate of $1e^{-4}$ using ADAM optmizer~\cite{kingma2017adam}.
The model is trained at an image resolution of 1620×3072 on random
crops of size 810 × 1536.
The color $\mathcal{L}_{1}$, structural $\mathcal{L}_\mathrm{ssim}$, and perceptual losses $\mathcal{L}_\mathrm{mrf}$, texel-level correspondence loss $\mathcal{L}_\mathrm{corr-tex}$ are assigned weights as $0.8$, $0.2$, $0.001$, and $0.2$, respectively. 

\section{Texel Super resolution} \label{suppsec:7}
\par \noindent \textbf{Network Architecture.} The texel super-resolution module predicts the residuals of the animatable Gaussian textures generated by the geometry and appearance networks.
It takes the channel-wise concatenated geometry and appearance textures as inputs and regresses their residuals to refine both geometry and appearance attributes.
Following the design of the geometry and appearance network, we employ a UNet-style architecture: the Gaussian texture attributes are first encoded and fused through two convolutional layers with 64 channels.
Subsequently, a convolutional layer processes the fused features to regress Gaussian textures at doubled original resolution.

\par \noindent \textbf{Training Details.} We train the texel super-resolution module with a combined loss of the color $\mathcal{L}_{1}$, structural $\mathcal{L}_\mathrm{ssim}$, and perceptual losses $\mathcal{L}_\mathrm{mrf}$, weighted as $0.8$, $0.2$, $0.001$, respectively.
The training lasts for 1,000,000 iterations with a learning rate of $1e^{-4}$ using ADAM optimizer~\cite{kingma2017adam}.
%

%
%
\section{Applications} \label{suppsec:8}
In this section, we will introduce texture editing enabled by \titleabr.

\par

\noindent \textbf{Texture Editing.} As discussed in the main paper, \titleabr produces ultra-detailed geometry with consistent triangulation and enhanced correspondence, enabling various applications—including consistent texture editing.
In Fig.~\ref{fig:suppltexture}, we show the results of texture editing performed through the following steps:
We first select an image with an alpha channel to serve as the texture for the surface geometry.
This texture is then rasterized onto the mesh, producing corresponding color and alpha images.
Finally, we apply alpha blending to combine the rasterized output with the rendered Gaussian Splat images.

As shown in Fig.~\ref{fig:suppltexture}, the inserted texture deforms consistently with the clothing wrinkles and remains firmly anchored to the character’s original texture.
This underscores the precision of both the reconstructed surface geometry and its correspondence over time, made possible by our multi-level surface alignment design.

\section{Qualitative Comparison} \label{suppsec:9}
\begin{figure*}[h]
    \centering
    \includegraphics[width=\linewidth]{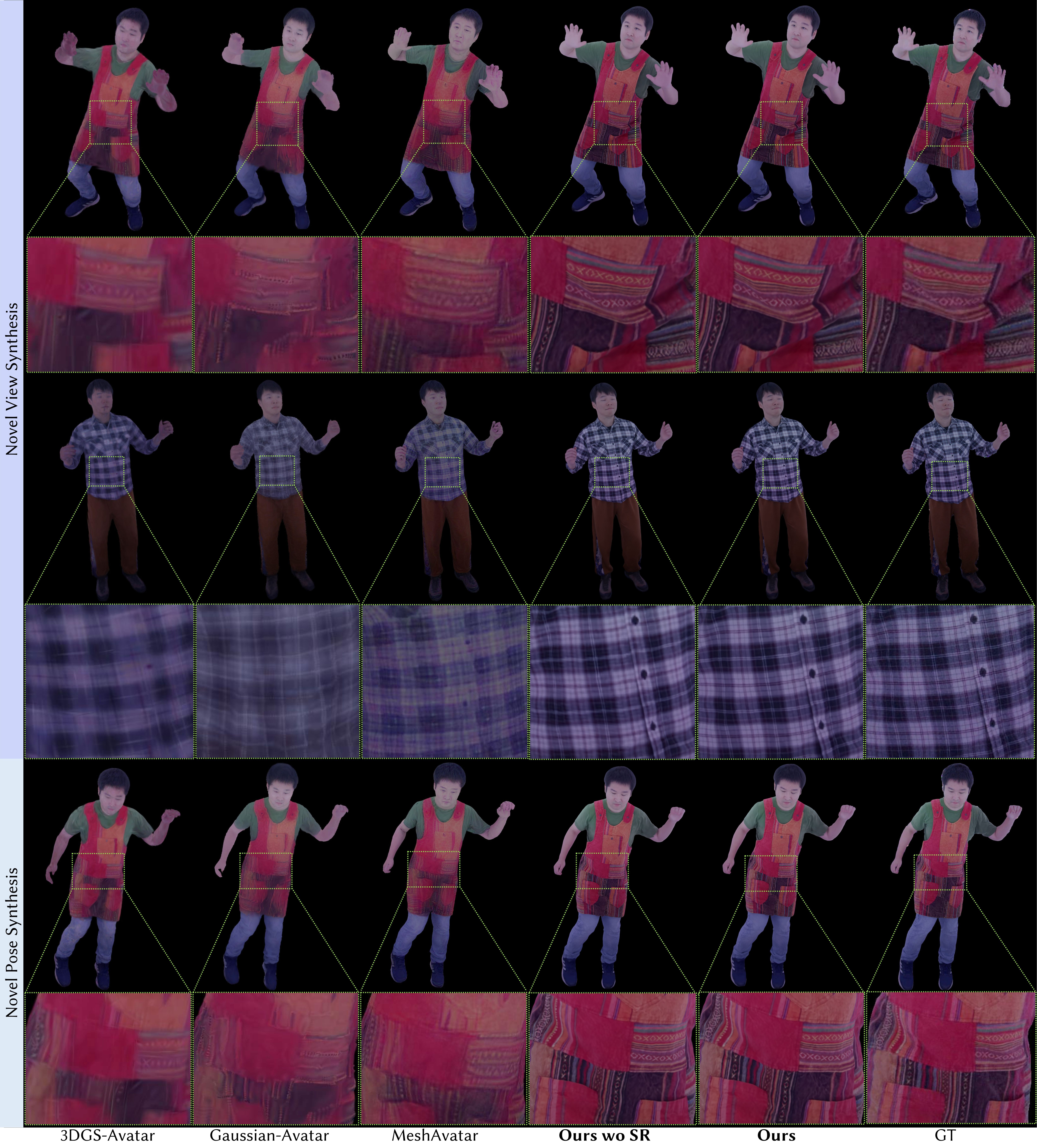}
    \vspace{-2em}
    \caption{
    \textbf{Qualitative Rendering Comparison.} We present more comparisons between approach with the competing approaches on novel view synthesis and novel pose generation.
    Please \textbf{zoom-in} to better observe the details.
    }
    \label{figsupp:qualcomparison}
\end{figure*}

In Fig.~\ref{figsupp:qualcomparison}, we provide additional qualitative comparison with the state of the art approaches on novel-view and novel pose synthesis task.
Thanks to the depth, vertex, and texel-level alignment, our approach (\textbf{Ours wo SR}) already achieves substantially improved fidelity, particularly in recovering fine-grained appearance details.
By introducing the texels super-resolution, the rendering quality is further enhanced,  for instance, the grid patterns on the shirt exhibit noticeably sharper borders.

%
%

\end{document}